\documentclass[fleqn]{article}

\usepackage{arxiv}

\usepackage{mypackages}
\bibliography{MyRefGlobal}
\usepackage{mycommands}

\usepackage[utf8]{inputenc} 
\usepackage[T1]{fontenc}    
\usepackage{hyperref}       
\usepackage{url}            
\usepackage{booktabs}       
\usepackage{amsfonts}       
\usepackage{nicefrac}       
\usepackage{microtype}      
\usepackage{cleveref}       
\usepackage{lipsum}         
\usepackage{graphicx}
\usepackage{doi}

\usepackage{float}

\title{Bayesian Statistical Modeling with Predictors from LLMs}

\date{}

\newif\ifuniqueAffiliation
\uniqueAffiliationtrue

\ifuniqueAffiliation 
  \author{ Michael Franke\thanks{Corresponding author.} \\
	Department of Linguistics\\
	University of Tübingen\\
	\texttt{mchfranke@gmail.com} \\
	\And
	Polina Tsvilodub \\
	Department of Linguistics\\
	University of Tübingen\\
	\texttt{polina.tsvilodub@gmail.com} \\
	\And
	Fausto Carcassi \\
	ILLC\\
	University of Amsterdam\\
	\texttt{fausto.carcassi@gmail.com} \\
}
\else
\usepackage{authblk}

\author{Michael Franke, Polina Tsvilodub, Fausto Carcassi}
\affil{Department of Linguistics\\University of Tübingen\\
\texttt{[michael.franke|polina.tsvilodub|fausto.carcassi]@uni-tuebingen.de}}
\fi

\renewcommand{\headeright}{}
\renewcommand{\undertitle}{}
\renewcommand{\shorttitle}{Bayesian Statistical Modeling with Predictors from LLMs}

\hypersetup{
pdftitle={Statistical modeling with LLM predictors},
pdfsubject={},
pdfauthor={Michael Franke, Polina Tsvilodub, Fausto Carcassi},
pdfkeywords={statistical models, language use, large language models, Bayesian data analysis, reference games},
}

\begin{document}
\maketitle

\begin{abstract}
  State of the art large language models (LLMs) have shown impressive performance on a variety of benchmark tasks and are increasingly used as components in larger applications, where LLM-based predictions serve as proxies for human judgements or decision.
  This raises questions about the human-likeness of LLM-derived information, alignment with human intuition, and whether LLMs could possibly be considered (parts of) explanatory models of (aspects of) human cognition or language use.
  To shed more light on these issues, we here investigate the human-likeness of LLMs' predictions for multiple-choice decision tasks from the perspective of Bayesian statistical modeling.
  Using human data from a forced-choice experiment on pragmatic language use, we find that LLMs do not capture the variance in the human data at the item-level.
  We suggest different ways of deriving full distributional predictions from LLMs for aggregate, condition-level data, and find that some, but not all ways of obtaining condition-level predictions yield adequate fits to human data.
  These results suggests that assessment of LLM performance depends strongly on seemingly subtle choices in methodology, and that LLMs are at best predictors of human behavior at the aggregate, condition-level, for which they are, however, not designed to, or usually used to, make predictions in the first place.
\end{abstract}


\section{Introduction}
\label{sec:introduction}

Enabled by the invention of deep neural transformer architectures \citep{VaswaniShazeer2017:Attention-is-Al}, recent years have brought a new generation of powerful large language models  \citep{DevlinChang2019:BERT:-Pre-train,ChungHou2022:Scaling-Instruc,OpenAI2023:GPT-4-Technical,TouvronLavril2023:LLaMA:-Open-and}.
State-of-the-art LLMs excel on many benchmark data sets \citep[e.g.,][]{srivastava2023-BIGbench,PerezRinger2023:Discovering-Lan}, and so promise to serve as foundation models for a vast and diverse set of applications, both in industry and academia \citep{BommasaniHudson2021:On-the-opportun}.
Yet, for any downstream application of LLMs, it is crucial to understand what LLMs can or cannot reliably do.

The way in which LLM capabilities should be assessed depends on what their intended application is.
For many industrial applications, the prevalent approach towards characterizing the capabilities of LLMs relies on benchmark testing, which usually consists in assessing the accuracy of LLM predictions in tasks where a designated ``target answer'' or ``gold standard'' exists, averaged over many instances of this task \citep[e.g.,][]{srivastava2023-BIGbench}, but more encompassing approaches also highlight the importance of more holistic assessments of LLM, including factors such as robustness, fairness and efficiency \citep{LiangBommasani2023:Holistic-Evalua}.
Benchmark-driven assessments are very useful for engineering purposes, when the main issue is whether a given system can perform a particular task correctly.



There are also applications of LLMs where benchmark testing on a ``gold standard'' is arguably not optimal.
Recent works increasingly go beyond using LLMs based on single-run input-output behavior, and instead utilize LLMs as a part of a larger computational process.
Simple examples include sophisticated prompting strategies \citep[e.g.,][]{LiuLiu2022:Generated-Knowl}, or structured reasoning models \citep[e.g.,][]{CreswellShanahan2022:Selection-Infer,GaoMadaan2023:PAL:-Program-ai,ParanjapeLundberg2023:ART:-Automatic-,YangKlein2023:DOC:-Improving-}.
More sophisticated examples include neuro-symbolic models in which LLMs supply specific parts of the relevant information for some practical application \citep[e.g.,][]{NyeTessler2021:Improving-Coher}, or where LLMs are a part of bigger programs to build towards something more akin to explanatory cognitive models \citep[e.g.,][]{WongGrand2023:From-Word-Model}.
For example, information from LLMs can be used to generate alternatives for deliberation \citep[e.g.][]{LewTessler2020:Leveraging-Unst,TsvilodubCarcassi2024:Towards-Neuro-S}, arguably similar to human resource-rational reasoning in open-ended domains \citep{VulGoodman2014:One-and-Done-Op}.
LLMs may also be used to rank or numerically score options in large or open-ended applications, e.g., to mimic human judgements of desirability, relevance or interestingness \citep[e.g.,][]{BaiKadavath2022:Constitutional-,ParkOBrien2023:Generative-Agen,KwonXie2023:Reward-Design-w,ZhangLehman2023:OMNI:-Open-ende}.
In many of these applications, LLMs essentially serve as a cheap, compressed stand-in for (average) human judgements, associations or choice behavior.
To assess the quality of LLMs in such contexts, it is less important to compare against a gold standard, and more important to compare against the full distribution of the human behavior that is to be captured by the LLM.
In sum, at least for some practical applications, the ``gold standard'' is not a single ``correct'' answer, but the full distribution of human responses.

Taking inspiration from experimental psychology, an increasing number of studies compares LLM predictions to human choice behavior in psychological experiments and investigates whether LLMs predict patterns of human answer behavior  \citep[e.g.,][]{BinzSchulz2023:Using-cognitive,Hagendorff2023:Machine-Psychol,ShiffrinMitchell2023:Probing-the-psy}.
The main focus is often to compare qualitative patterns in LLM predictions and human data, but there is also work investigating whether LLMs can make adequate \emph{quantitative predictions}.
Most notably, there is a strong tradition of relatively early work in computational psycholinguistics \citep{MarvinLinzen2018:Targeted-Syntac,HuGauthier2020:A-Systematic-As}, which investigates whether quantitative predictions derived from language models match quantitative aspects in human experimental data, such as reading times \citep{WilcoxVani2021:A-Targeted-Asse} or the amplitude of the N400 component of event-related-potentials in EEG measurements \citep{LindborgRabovsky2021:Meaning-in-brai}.

The work presented in this paper seeks to extend the investigation of the human-likeness of predictions derived from LLMs.
Our foremost concern is methodological: How can we derive full distributional predictions from information supplied by LLMs, and how can we stringently test whether these distributional predictions are adequate given suitable empirical data?
In short, while a lot of previous like-minded work has used targeted assessment of LLM capabilities from the point of view of the experimental psychologist, we here adopt the more specific perspective of the probabilistic / statistical modeller.
Taking numerical predictor values generated from LLMs as input, we explore strategies of building a Bayesian statistical model around them, and to scrutinize these LLM-grounded Bayesian statistical models with the usual methods of Bayesian data analysis, in particular model criticism \citep{GelmanCarlin2014:Bayesian-Data-A}.

\begin{figure}[t]
  \centering
  \includegraphics[width = 0.95\textwidth]{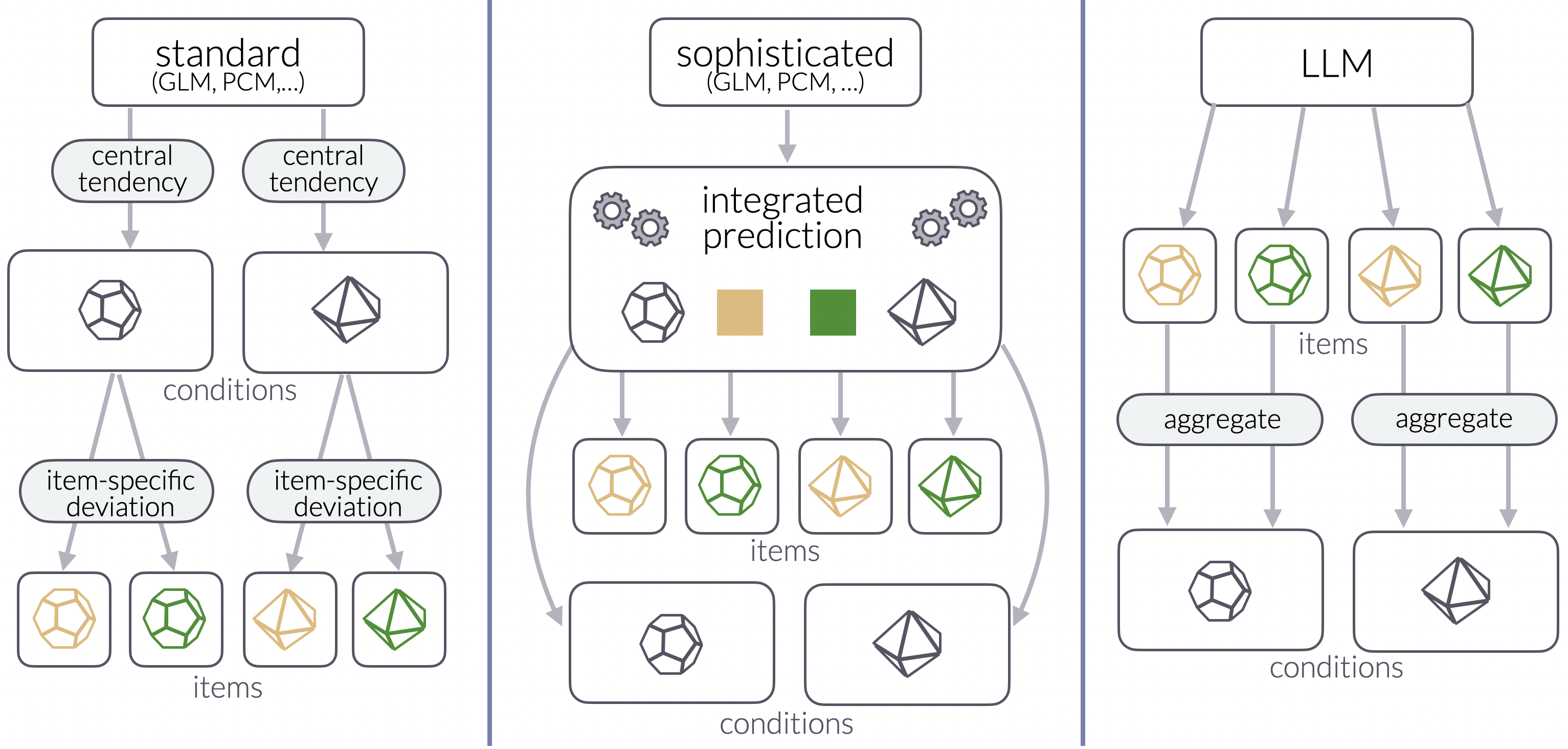}
  \caption{
    Schematic representation of key conceptual differences between different types of predictive models.
    Standard statistical models, like hierarchical regression models, typically make predictions for aggregate data (e.g., at the condition level), and add random offsets for item-level variation.
    More sophisticated models, like some probabilistic cognitive models, may holistically combine information from an aggregate level (task, condition) with specific information about items.
    LLMs, in contrast, first and foremost give prediction about each individual item and must rely on proper aggregation to arrive at condition-level predictions.
  }
  \label{fig:stats-model-types}
\end{figure}

A main conceptual take-away of this investigation, is that statistical models built around LLMs are, by design, fundamentally different from common statistical or probabilistic cognitive models; an observation which also reflects back on the possibility of seeing LLMs as potentially explanatory model for human behavior or cognition.
Concretely, LLMs make predictions for each individual item, rather than specifying a predictor of central tendency at a more aggregate level, such as at the level of an experimental condition, as standard statistical models or probabilistic cognitive models normally do (see Figure~\ref{fig:stats-model-types}).
As this point is crucial for our investigation, the following elaborates briefly.

Psychological research into the workings of the human mind aims to find generalizable patterns in the way information is processed within or across different domains of cognition.
Experimental work therefore often compares human performance in different \emph{experimental conditions}, which reflect the general factors that are hypothesized to influence behavior.
Yet a single experimental condition is often instantiated with different \emph{experimental items}, which are usually not under scrutiny for any systematic, predictable effect on the observed measurements.
For example, classical research on human memory \citep{AtkinsonShiffrin1968:Human-memory:-A} investigated the effect of rehearsal on memory consolidation.
Relevant experiments compare recall with and without rehearsal (experimental conditions), while using different items (words, numbers, etc., to be memorized) in each instantiation of the same memory task.
Likewise, when studying how hearing a color word can facilitate a same-or-different judgement of color swatches \citep{Rosch1975:The-Nature-of-M}, the main experimental manipulation concerns the typicality of shown color swatches, while the variability between different color words like ``blue'' or ``green'', is less important to this research question and so treated as \emph{item-level variation}.
Consequently, a typical psychological experiment is mainly interested in assessing behavior at the level of the experimental condition, because that is where the distinctions relevant to the research question reside.
Nevertheless, each experimental condition can be, and often is, instantiated with different items, variation among which is deemed less relevant to the research question at hand.

Data from human participants for experiments of this kind usually show some variability between items, and also variability between participants.
This variability is commonly incorporated in standard statistical models as random stochastic variation, e.g., by using hierarchical regression models \citep{Jaeger2008:Categorical-dat,barr2013,SorensenHohensteinb2016:Bayesian-linear}, as shown on the left side of Figure~\ref{fig:stats-model-types}.
Still, the focus of interest usually remains at the condition-level effects, because it is this more abstract level of behavioral aggregation that is relevant for generalizable theory building.
Similarly, when analyzing or explaining data from psychological experiments with a probabilistic cognitive model (PCM), the model's predictions will naturally be set-up to predict data by taking condition-level properties, possibly in conjunction with item-level properties, into account \citep[e.g.,][]{NilsonRieskamp2011:Hierarchical-Ba,Lee2011:How-Cognitive-M,ScheibehenneRieskamp2013:Testing-the-Ada}, as shown schematically in the middle of Figure~\ref{fig:stats-model-types}.
On the other hand, LLMs first and foremost provide predictions about each item.
While it may be the case that, in producing an item-level prediction, the internal computation of a powerful LLM is informed by computations that incorporate abstract knowledge roughly corresponding to the condition-level, the atomic predictions accessible to the common user are specific to each individual string tested.
Consequently, this points to known concerns about robustness of predictions under perturbations of input prompts \citep[e.g.][]{ReynoldsMcDonell2021:Prompt-Programm,WebsonPavlick2022:Do-Prompt-Based,SalinasMorstatter2024:The-Butterfly-E,TsvilodubWang2024:Predictions-fro}.

These considerations raise two important conceptual and empirical questions.
First, we need to ask whether the item-level predictions made by LLMs are empirically correct, i.e., match the human data at the item-level.
Second, the implied methodological challenge lies in specifying how item-level information can be used, e.g., by different aggregation methods, to make robust predictions at a more abstract level (condition, task, $\dots$).
To investigate these issues, this paper introduces different ways of building a Bayesian probabilistic model around core predictor values derived from various LLMs.
We then use the standard tools of Bayesian data analysis to fit and check the resulting statistical models based on data from human multiple-choice experiments.
Rather than aiming for scale and large-coverage, we focus on transparency and zoom in on a single case study of pragmatic language production and interpretation, namely so-called reference games \citep[e.g.,][]{DeemterSluis2006:Building-a-Sema,DegenFranke2013:Cost-Based-Prag,QingFranke2013:Variations-on-a,Frank2016:Rational-speech,GrafDegen2016:Animal-dog-or-d,SikosVenhuizen2021:Reevaluating-pr}.
The Rational Speech Act framework \citep{FrankGoodman2012:Predicting-Prag} provides a widely used probabilistic model for human data from reference games, so that we can compare a theoretically motivated probabilistic cognitive model (PCM) against a probabilistic model built on top of LLM predictions.
We find that variability predicted by the LLM at the item-level is generally not borne out by the human data, that not all ways of constructing condition-level predictions are equally good, and that different LLMs as backends may prefer to use different aggregation strategies.

The paper is structured as follows.
Section~\ref{experiment-reference-games} describes an experiment with human participants with a text-based reference game.
Section~\ref{sec:model-pred-from} introduces the Rational Speech Act (RSA) model for the human data from the reference game experiment.
Section~\ref{sec:item-level-pred} exposes a statistical model for item-level predictions derived from LLM scores and investigates whether this adequately captures the human data at the item-level based on scores from GPT-3.5 \citep{BrownMann2020:Language-Models}.
Section~\ref{llm-predictions-for-reference-games} discusses different ways of deriving probabilistic predictions from LLMs at the condition-level and compares them against the human data and each other.
Finally, Section~\ref{sec:gener-other-llm} explores whether previous results generalize to other LLM backends by investigating different versions of LLaMA2 \citep{TouvronLavril2023:LLaMA:-Open-and}.\footnote{
  All data and code is available at this OSF repository:  \href{https://osf.io/f6j3a/?view_only=5e820cc8bbee4549aed58dc252ba61b9}{https://osf.io/f6j3a/?view\_only=5e820cc8bbee4549aed58dc252ba61b9}.
}

\section{experiment: Reference games}
\label{experiment-reference-games}

\begin{figure}
  \centering

  \includegraphics[width = 0.8\textwidth]{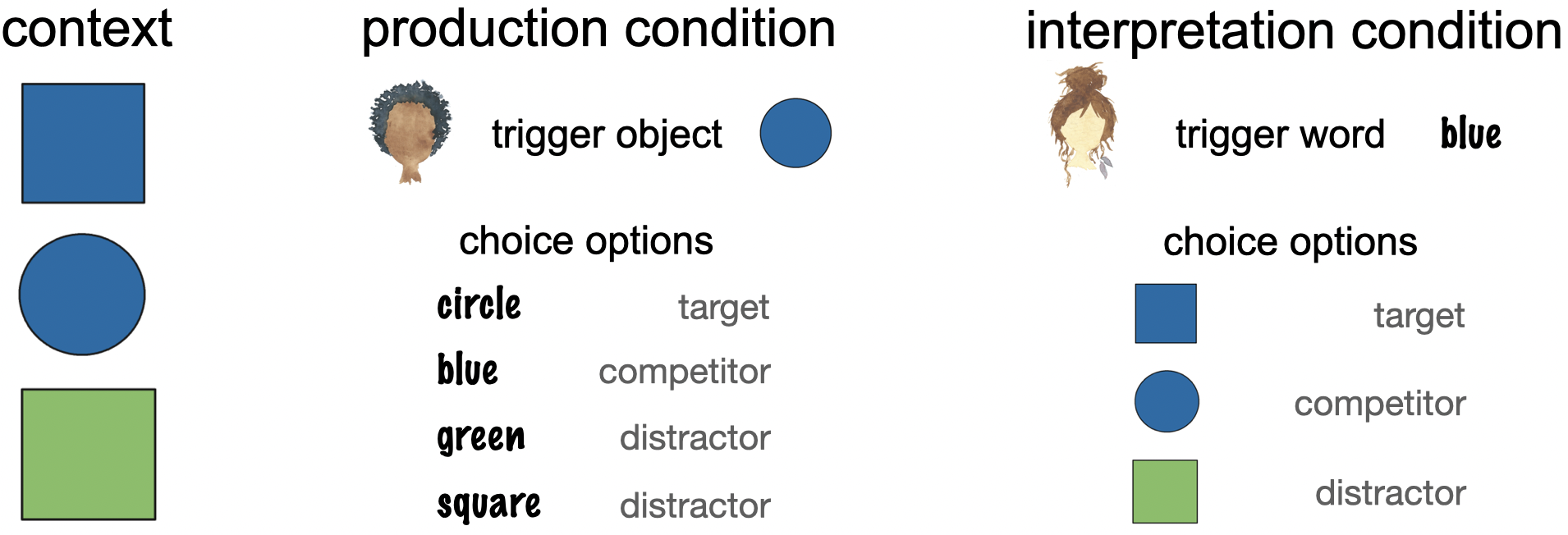}

  \caption{Structure of a reference game with human participants. Each trial consists of a set of objects, the so-called context. In production trials, participants choose a single word to describe a trigger object from the context. In interpretation trials, an object is selected as the likely object a trigger word is referring to.}
  \label{fig:ref-game}
\end{figure}

Reference games are an established, well-understood and austere experimental paradigm to test human decision making in abstract communicative tasks.
A reference game consists of two players, a speaker and an interpreter, who jointly observe a set of objects, usually referred to as context (see Figure~\ref{fig:ref-game}).
In the \textbf{production condition}, the speaker is assigned a \emph{trigger object} from the context set which they have to describe to the interpreter.
In the \textbf{interpretation condition}, the interpreter observes a description, here called \emph{trigger word}, and chooses one of the objects from the context set.
The goal of the game is, for the speaker, to choose a description that enables the interpreter to choose the trigger object; and, for the interpreter, to guess correctly which object the speaker had in mind when using the trigger word.

The example in Figure~\ref{fig:ref-game} is a standard case, which we will use throughout, where human choices are informative about the pragmatic reasoning that human decision makers engage in.
In this example, there are two features that differ across three objects (here shape and color).
One object shares both its color and shape with one other object, while the two other objects have one unique feature (e.g., being the only circle, or the only green object).
In a critical production trial, the trigger object to describe is one of the two objects with a unique feature.
The speaker has four words to choose from.
The \textbf{target utterance} is the word which uniquely describes the trigger object.
The \textbf{competitor utterance} is the word that is true of the trigger object, but also true of another object.
The other utterances, both of which are false of the trigger object are \textbf{distractor utterance}.
In a critical interpretation trial, the trigger word is the one that is true of two of the three objects.
If participants engage in pragmatic thought, they might reason that \emph{if} the speaker had wanted to refer to one of the two objects of which the trigger word is true (blue square and blue circle in Figure~\ref{fig:ref-game}), the speaker could have used a more informative word for exactly one of those two objects (``circle''), so they are more likely to refer to the \textbf{target object} (the blue square in Figure~\ref{fig:ref-game}).
The \textbf{competitor object} is the other object of which the trigger word is true.
The \textbf{distractor object} is the object of which the trigger word is false.

We implemented a simple reference game for human participants in which each trial instantiated the structure of the example shown in Figure~\ref{fig:ref-game}.
While previous reference games with human participants used pictorial representations of objects, and sometimes even pictorial representations of messages, we implemented a text-only version in order to be able to compare the predictions of LLMs for human data, when both LLMs and humans processed the same textual representation of the stimuli.
The experiment was realized as an online task using \texttt{magpie} \citep{FrankeJi:magpie:-Minimal}.\footnote{
  The code for the experiment can be found at \href{https://github.com/magpie-ea/magpie3-text-refgame}{https://github.com/magpie-ea/magpie3-text-refgame}, and a live version of the experiment can be tested at \href{https://magpie-ea.github.io/magpie3-text-refgame/}{https://magpie-ea.github.io/magpie3-text-refgame/}.
}

\paragraph{Participants.}
A total of 302 participants were recruited via Prolific for monetary
compensation (\textsterling0.45, corresponding to roughly \textsterling 15.40 per hour).
All participants self-identified as native speakers of English.

\paragraph{Materials \& design.}
We created 100 different items as stimulus material via a stochastic process.
Each item is a different textual description of a reference game with the same logical structure as the example from Figure~\ref{fig:ref-game}.
For each item, the context consists of three objects.
As in the original paper by \citet{FrankGoodman2012:Predicting-Prag}, objects are defined by a triple of properties, namely a color, a shape and a texture.
For each property, there were four possible values, e.g., blue, green, red, and orange for color.
The sampled items differed in terms of the properties of the objects in the context set, and in terms of the order in which the objects and expression alternatives were presented in the text.
Figures~\ref{fig:refgame-screenshot-production} and \ref{fig:refgame-screenshot-interpretation} from Appendix~\ref{sec:scre-from-online} show example screenshots from the experiment.

\paragraph{Procedure.}
Each participant saw four different items sampled randomly from the pre-generated item set.
Participants first played two of these in the production condition, then the other two in the interpretation condition.

\paragraph{Results.}
The overall distribution of choices that correspond to the target, competitor, and distractor states is shown in Figure~\ref{fig:refgame-counts} (together with model checking results to be introduced later).\footnote{
  The production condition actually has two distractor choices.
  Here and in the following, these are lumped together as a single category, also when modeling random errors in later models.
  }
It is interesting that the distractor options were chosen rather often.
We also see that the number of target choices is higher in the production condition than in the interpretation condition.
This is in line with previous experimental results on human reference games.
For example, in previous forced-choice reference games with human participants with pictorial presentations of objects, \citet{QingFranke2013:Variations-on-a} observed the following proportions of target, competitor and distractor options: $\tuple{0.882, 0.118, 0}$ in the production and $\tuple{0.492, 0.506, 0.003}$ in the interpretation condition (for 288 observations in each condition).

\section{Model predictions from probabilistic pragmatics}
\label{sec:model-pred-from}

\begin{figure}[t]
  \centering
  \includegraphics[width = 0.9 \textwidth]{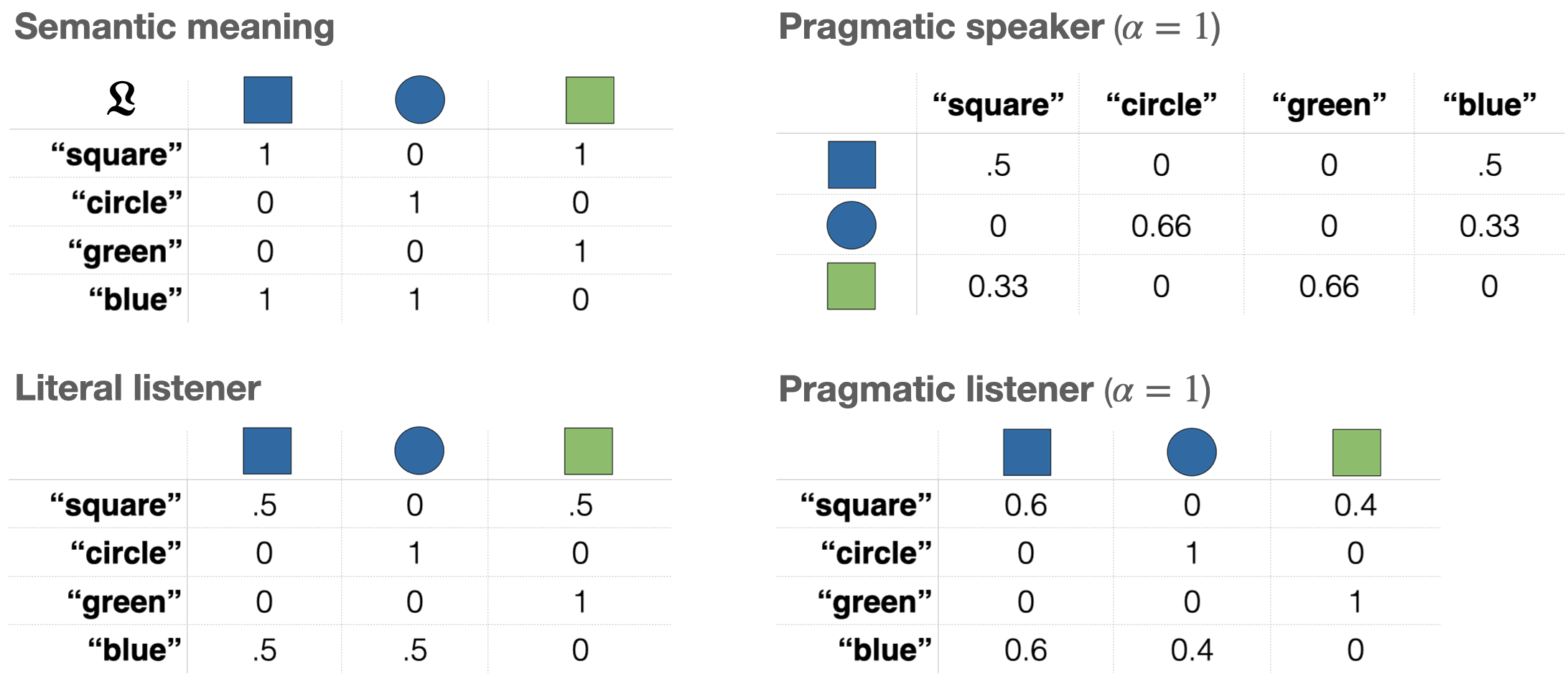}
  \caption{
    Example of predictions from the RSA model.
    The semantic meaning is shown as a matrix of binary truth-values.
    The policies of literal listener, pragmatic speaker and listener are calculated for uniform priors over states (referents) for $\alpha=1$, and are shown as row-stochastic matrices.
  }
  \label{fig:RSA-example}
\end{figure}

Data from reference games with human participants have been variously analyzed with probabilistic models using inspiration from behavioral game theory \citep[e.g.,][]{DegenFranke2013:Cost-Based-Prag}, probabilistic Bayesian modeling \citep[e.g.,][]{FrankGoodman2012:Predicting-Prag} or other forms of probabilistic modeling \citep[e.g.,][]{GattGompel2013:Are-we-Bayesian}.
Common to these approaches is that they derive or define, based on some explicit conceptual motivation, a parameterized stochastic speaker policy, $P_{S}(u \mid s; \theta_{S})$, modulated by parameters $\theta_{S}$, for a speaker's choice of expression or utterance $u$ given a referent or state $s$, which the speaker wants to communicate;
and a parameterized stochastic listener policy, $P_{L}(s \mid u; \theta_{L})$, capturing the probability of choosing a referent $s$ for utterance $u$.

\paragraph{The Rational Speech Act model.}
As a concrete example, we introduce the Rational Speech Act (RSA) model first described in this form by \citet{FrankGoodman2012:Predicting-Prag} \citep[for overview see][]{FrankeJager2015:Probabilistic-p,GoodmanFrank2016:Pragmatic-Langu,StevensBenz2018:Game-Theoretic-,ScontrasTessler2021:A-practical-int,Degen2023:The-Rational-Sp}.
The RSA model defines pragmatic reasoning as a sequence of iterated (soft-)optmization of policies and Bayesian inference, grounding out in literal interpretation.
If $\mathfrak{L}(s,u) \mapsto \set{0,1}$ is a semantic meaning function mapping each pair of state $s$ and utterance $u$ to a (binary) truth-value, and if $P_{\text{prior}}(s)$ is a prior over states, a literal listener policy is defined as:
\begin{align*}
 P_{L_{0}}(s \mid u) \propto \mathfrak{L}(s,u)\  P_{\text{prior}}(s)\,.
\end{align*}
The pragmatic speaker policy is defined as soft-optimizing the choice of utterance to minimize the literal listener's surprisal for the state to be communicated, i.e., to maximize the log-probability of the trigger object given the utterance:
\begin{align}
  \label{eq:RSA-speaker}
  P_{S}(u \mid s, \alpha) & \propto \expo \left [ \log P_{L_{0}}(s \mid u) \right ] \,.
\end{align}
Finally, the pragmatic listener is defined as the policy resulting from applying Bayes rule, solving the inverse-problem for the previously defined speaker policy:
\begin{align}
  \label{eq:RSA-interpreter}
  P_{L}(s \mid u, \alpha) \propto  P_{S}(u \mid s, \alpha) \  P_{\text{prior}}(s) \,.
\end{align}

Figure~\ref{fig:RSA-example} gives example calculations (assuming a flat prior and $\alpha=1$) for the reference game from Figure~\ref{fig:ref-game}.
For $\alpha=1$, the model predicts that the probabilities of target, competitor and distractor options are $\tuple{\nicefrac{2}{3}, \nicefrac{1}{3}, 0}$ for the production, and $\tuple{\nicefrac{3}{5}, \nicefrac{2}{5}, 0}$ for the interpretation condition.
Increasing $\alpha$ will increase the odds of target over competitor choices.

\paragraph{Condition-level predictions.}
In sum, the condition-level predictions of the RSA model are a parameterized function $P_{\text{cond}}^{\text{RSA}}(R_{l}, C; \alpha_{c})$, assigning a probability to each response category $R_{l}$ (target, competitor, or distractor) in each condition $C$ (production or interpretation) for a given $\alpha_{c}$.

\begin{figure}[t]
  \centering

    \includegraphics[width=0.9\linewidth]{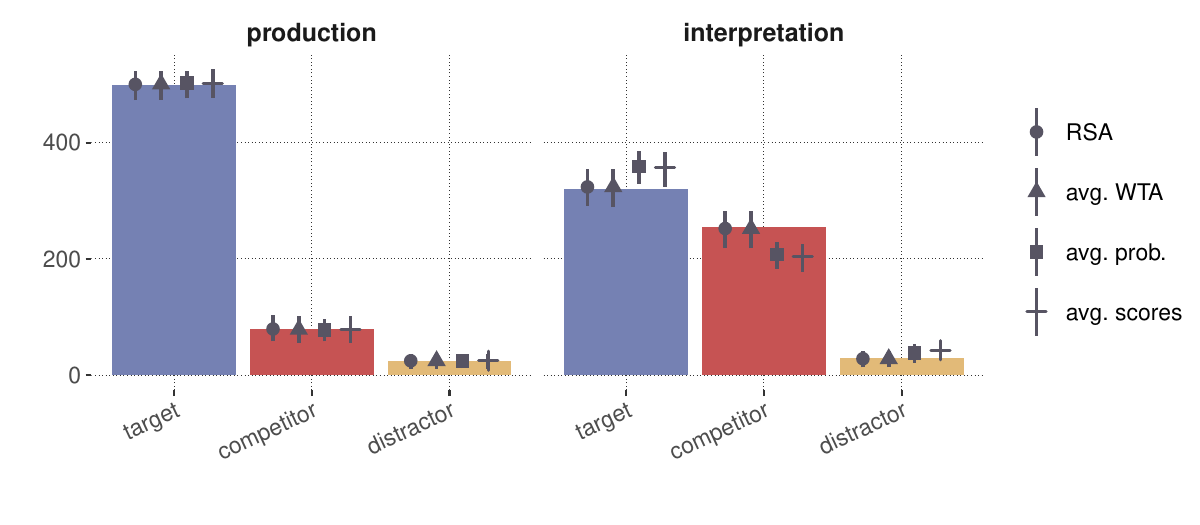}

    \caption{
      Counts of choices from reference games with human participants (colored bars), with summary statistics from the posterior predictive distribution of four models (shapes and error bars).
      Shapes show the mean of the posterior predictive distributions of the RSA model and three aggregated condition-level predictors derived from item-level LLM scores (introduced in Section~\ref{llm-predictions-for-reference-games}).
      Error-bars show corresponding 95\% credible intervals of the posterior predictive.
    }
  \label{fig:refgame-counts}
\end{figure}

The model constructed so far predicts probability zero for distractor choices, so that the human data shown in Figure~\ref{fig:refgame-counts}, where the distractor option was chosen in both conditions, would immediately rule out the model entirely.
It is therefore common to include a small error probability $\epsilon$, with which a choice would be made at random \citep[e.g.,][]{LeeWagenmakers2013:Bayesian-Cognit}, so that we get:
\begin{align*}
  P_{\text{cond}}^{\text{RSA}}(R_{l}, C; \alpha_{c}, \epsilon_{c}) = (1 - \epsilon_{c}) \  P_{r}(R_{l}, C; \alpha_{c}) +  \nicefrac{\epsilon_{c}}{3}\,,
\end{align*}
where $\epsilon_{c}$ is a (condition-specific) parameter giving the probability that a choice was made by randomly guessing.\footnote{
  Since the RSA model predicts probability 0 for the distractor option, this model is, in principle, able to predict any probability distribution over the three choice categories that is compatible with the order: $P_{r}(R_{t}) \ge P_{r}(R_{c}) \ge P_{r}(R_{d})$.
  Intuitively, this is because with $\epsilon=0$, $P_{r}(R_{d}, C; \alpha_{c}) = 0$, so that there is an $\alpha$ for any ratio of predicted choice probabilities for target and competitor, as long as the target probability is no smaller than the competitor probability.
  The $\epsilon$-transformation is essentially a linear shift in the probability simplex towards the maximum entropy prediction, so that every prediction which obeys the ordering restriction above can be made for some pair of $\alpha$ and $\epsilon$.
  This prediction-triviality is met in two ways.
  For one, the Bayesian priors on model parameters soft-constrain the model, so that the \emph{ex ante} credible predictions do rule out many logically possible observations.
  For another, we break the triviality by assigning a non-zero probability to the prediction for the distractor option.
  The same triviality problem lurks for the \emph{average-WTA} model introduced in Section~\ref{llm-predictions-for-reference-games}, and the same solution is applied to it.
}

The data $D_{C}$ from condition $C$, see Figure~\ref{fig:refgame-counts}, consists of counts for each response category.
The parameterized likelihood function entailed by the RSA model for condition-level data $D_{C}$ is:
\begin{align}
  \label{eq:RSA-likelihood}
 P^{\text{RSA}}_{\text{cond}}(D_{C} \mid C, \alpha_{C}, \epsilon_{C}) = \text{Multinomial}(D_{C}, \tuple{P_{\text{cond}}^{\text{RSA}}(R_{l}, C; \alpha_{c}, \epsilon_{c}}_{1 \le l \le 3})\,.
\end{align}
The result is a four-parameter model, one pair of parameters per condition.

\paragraph{Bayesian posteriors \& model checking.}
Parameterized predictions, like in Equation~(\ref{eq:RSA-likelihood}), can be assessed in the light of the empirical data with the usual tools of Bayesian data analysis \citep[e.g.][]{GelmanCarlin2014:Bayesian-Data-A,McElreath2016:Statistical-Ret,Lambert2018:A-Students-Guid}.
Let $\alpha_{c}\sim \text{log-Normal}(1,1)$ have a reasonably wide log-Normal prior, and let $\epsilon_{c} \sim \text{Beta}(1,15)$ have a Beta prior favoring small values.
Using Stan \citep{Team2023:The-Stan-Core-L} for Bayesian inference, we obtain estimates of posterior credible values of model parameters (summary statistics of which are shown in Table~\ref{tab:sumStats}).\footnote{
  Posterior samples where generated for four chains with 2000 samples each, after a warm-up of 1000 samples. The ``adapt-delta'' value was set to 0.99. Converge was checked with $\hat{R}$-statistics \citep{GelmanRubin1992:Inference-from-}.
}

\begin{table}[t]
\centering
\begin{tabular}{cllllcrlcrlc}
  \toprule
  &&&&&& $\alpha$ &&& $\epsilon$ & \\ \cmidrule(r){6-8} \cmidrule(l){9-11}
  & model & data & method & condition & |95\% & mean\ & 95\%| & |95\% & mean\ & 95\%| & Bpppv \\
  \midrule
  \textcolor{fern-main}{$\checkmark$} & RSA & item  & ---         & prd. & 2.62 & 3.13 & 3.69  & 0.08 & 0.12 & 0.16 & 0.29 \\
  \textcolor{fern-main}{$\checkmark$} & RSA & item  & ---         & int. & 0.21 & 0.67 & 1.08  & 0.08 & 0.14 & 0.19 & 0.21 \\
  \textcolor{fern-main}{$\checkmark$} & RSA & cond. & ---         & prd. & 2.62 & 3.14 & 3.70  & 0.08 & 0.12 & 0.17 & 0.50 \\
  \textcolor{fern-main}{$\checkmark$} & RSA & cond. & ---         & int. & 0.27 & 0.68 & 1.13  & 0.09 & 0.14 & 0.19 & 0.51 \\ \addlinespace[0.5em]
  \textcolor{shimmer-main}{$\times$} & GPT & item  & ---         & prd. & 0.40 & 0.51 & 0.62  & 0.07 & 0.12 & 0.17 & 0.00 \\
  \textcolor{shimmer-main}{$\times$} & GPT & item  & ---         & int. & 0.52 & 0.66 & 0.82  & 0.00 & 0.05 & 0.15 & 0.00 \\
  \textcolor{fern-main}{$\checkmark$} & GPT & cond. & avg. scores & prd. & 0.61 & 0.74 & 0.87  & 0.08 & 0.12 & 0.17 & 0.49 \\
  \textcolor{shimmer-main}{$\times$} & GPT & cond. & avg. scores & int. & 0.99 & 1.17 & 1.35  & 0.00 & 0.02 & 0.06 & 0.00 \\
  \textcolor{fern-main}{$\checkmark$} & GPT & cond. & avg. prob.  & prd. & 0.81 & 5.00 & 14.58 & 0.08 & 0.12 & 0.16 & 0.61 \\
  \textcolor{shimmer-main}{$\times$} & GPT & cond. & avg. prob.  & int. & 1.36 & 2.04 & 2.67  & 0.00 & 0.03 & 0.08 & 0.00 \\
  \textcolor{fern-main}{$\checkmark$} & GPT & cond. & avg. WTA    & prd. & 0.86 & 1.04 & 1.22  & 0.08 & 0.12 & 0.17 & 0.48 \\
  \textcolor{fern-main}{$\checkmark$} & GPT & cond. & avg. WTA    & int. & 0.16 & 0.43 & 0.69  & 0.06 & 0.13 & 0.19 & 0.50 \\
   \bottomrule \\
\end{tabular}
\caption{
  Summary statistics for model fits.
  Columns show the means and 95\% credible intervals for posterior samples for each parameter, as well as the Bayesian posterior predictive $p$-values for each model, data type (item- or condition-level data), aggregation method (for condition-level data), and condition.
  The final column shows the Bayesian posterior predictive $p$-values.
  The first columns marks whether the row's Bayesian posterior predictive $p$-values are higher / lower than $0.05$, flagging relative inability to capture the structure of the data.
}
\label{tab:sumStats}
\end{table}

To assess goodness-of-fit, we use the \emph{posterior predictive distribution}, i.e., the model's predictions about data of the same size and structure as the training data.
Figure~\ref{fig:refgame-counts} shows summary statistics (means and 95\% credible intervals) for the posterior predictive distribution of the RSA model (among other models for condition-level data, which are to be introduced later).
We see that for both conditions the RSA model passes a ``visual posterior predictive check'' \citep{Kruschke2011:Doing-Bayesian-}, which requires that the distribution of posterior predictions includes the observed choice rates for each answer option.
To corroborate the visual impression, Table~\ref{tab:sumStats} shows sample-based estimates of Bayesian posterior predictive $p$-values (Bppp values), using likelihood of the observed data as a test statistics.
These Bppp values approximate the probability that a model conditioned on observed data $D_{\text{obs}}$ predicts future data $D_{rep}$, of the same size and format as $D_{obs}$, that is at least as likely as the data $D_{\text{obs}}$ itself is (given the posterior predictive distribution).
Very small Bppp values indicate that the model might be inadequate for reproducing the data it was trained on, so to speak.
This is clearly not the case for the RSA model with Bppp values close to 0.5, as shown in Table~\ref{tab:sumStats}.
In sum, the condition-level predictions by the RSA model, a theoretically motivated PCM, are not discredited by the condition-level data.
Reversely, the RSA model seems to adequately capture the condition-level data.

\paragraph{Item-level predictions.}
To test item-level predictions, the whole data set for condition $C$ is chunked into item-level data, $D_{C} = \tuple{D^{1}_{C}, \dots, D^{m}_{C} }$, where $D_{C}^{k}$ is the data collected for the $k$-th out of $m$ items for condition $C$.
Probabilistic cognitive models, like the RSA model, may holistically combine condition- and item-level information in their predictions, as shown in the middle of Figure~\ref{fig:ref-game}, e.g., by incorporating further parameters to capture variance based on different item-classes, such as the empirically observed preference for selecting shape terms over color terms in reference games \citep[e.g.][]{QingFranke2013:Variations-on-a}.
But for present purposes, we simply use the same (condition-level) predictor to separately predict data from each item, so that the likelihood function for item-level data becomes:
\begin{align*}
 P^{\text{RSA}}_{\text{item}}(D_{C} \mid C, \alpha_{C}, \epsilon_{C}) = \prod_{k = 1}^{m} \text{Multinomial} \left (D_{C}^{k}, \tuple{P_{\text{cond}}^{\text{RSA}}(R_{l}, C; \alpha_{c}, \epsilon_{c}}_{1 \le l \le 3} \right )\,.
\end{align*}
Fitting the RSA model to the partitioned data set, we find estimates of Bppp values that do not discredit the model (see Table~\ref{tab:sumStats}).
This suggests that the item-level variation in the human data is not so pronounced as to provide strong evidence against the condition-level predictor from the RSA model.
In other words, the condition-level RSA predictions seem to adequately capture also the item-level data.

The following sections apply the same methods of Bayesian model criticism also to models built around predictor values from LLMs.
Section~\ref{sec:item-level-pred} first looks at the item-level data, before Section~\ref{llm-predictions-for-reference-games} investigates different ways for deriving condition-level predictions.

\section{Item-level predictions from LLMs}
\label{sec:item-level-pred}

An (autoregressive) LLM is designed to predict the next token given an input string.\footnote{
  Nothing of substance changes when a Bayesian statistical model is built around scores from a masked language model.
  We focus on autoregressive, left-to-right language modeling and next-token prediction for ease of reference, since all models tested here are autoregressive.
}
This is essentially an item-level prediction: next-token probabilities are specified for concrete strings after a concrete instance of a task, i.e., an item of the task, not for the task as such.
From these next-token probabilities, we can derive probabilities for multiple-choice answers for each item (this section) and for a condition (next section).

\paragraph{Notation.}
Let $\set{I_{1}, \dots, I_{m}}$ be $m$ be instances of the same task, or items belonging to the same (logical) condition in a behavioral experiment.
Each item $I_{k} = \tuple{x_{k}, \tuple{y_{kl}}_{1 \le l \le n}}$ consists of an input prompt $x_{k}$, which is a string of text, and $n$ choice options $\tuple{y_{kl}}$, all of which are strings as well, possibly composed of $|y_{kl}|$ tokens, $y_{kl} = w_{kl1}, \dots, w_{kl|y_{kl}|}$.
For simplicity of notation, we assume that the $l$-th choice option $y_{kl}$ for each item $k$ belongs to the same response category $R_{l}$.
For the case at hand these categories are: target, competitor, distractor, so that $y_{k1}$ always corresponds to the designated \emph{target option}, $R_{1}$.

\paragraph{LLM scores \& predictions.}
The most obvious \emph{item-level score} an (autoregressive) LLM provides for each choice option $y_{kl}$ is its log-probability:\footnote{
  Commonly used item-level scores include corrections for variable length of answer options \citep[e.g.,][]{BrownMann2020:Language-Models} or variation in base rate among answer options \citep[e.g.,][]{HoltzmanWest2021:Surface-Form-Co}.
  Following common practice in the current literature, we also use length-corrected, average log-probabilities as raw scores for all options.
  As most options have identical number of tokens, this only affects the numerical values of softmax-parameter $\alpha$, as introduced below.}
\begin{align*}
  \text{S}_{kl}
  =  \sum_{i=1}^{|y_{kl}|} \log P_{\text{LLM}} \left(w_{kli} \mid x_{k}, w_{kl1}, \dots, w_{kl(i-1)} \right)  \,.
\end{align*}

Based on each option's score, we can define the \emph{item-level prediction} of choosing option $y_{kl}$ in terms of soft-maximization as:
\begin{align*}
P_{\text{item}}^{\text{LLM}}\left ( y_{kl} \mid C; \alpha_{c} \right ) \propto \expo \left (\alpha_{c} \ \text{S}_{kl} \right)\,.
\end{align*}
Notice that the usual item-level prediction used in benchmark testing is actually a ``winner-takes-all'' strategy, which would choose any option that maximizes the score, but this is just a special case of the above for $\alpha_{C} \rightarrow \infty$.
Here, the $\alpha_{C}$ parameter corresponds to inverse temperature, so that $\alpha_{C} \rightarrow \infty$ corresponds to greedy sampling with temperature zero.
Like in the RSA model, we use an independent soft-max parameter for each condition.

\paragraph{Bayesian statistical modeling with LLM predictors.}
As mentioned previously, the items of the reference game experiment from Section~\ref{experiment-reference-games} differ in which levels of features (color, shape, texture) instantiate the structure of the task shown in Figure \ref{fig:ref-game}, as well as the order of presentation of objects and words.
We expect both human and LLM predictions to vary between different items: e.g., humans seem to have preferences for some features \citep[e.g.,][]{QingFranke2013:Variations-on-a}, such as over-production of informationally irrelevant material \citep[e.g.,][]{DaviesKatsos2010:Over-informativ,Rubio-Fernandez2019:Overinformative,DegenHawkins2020:When-redundancy}; and machines may be susceptible to the presentation of the order of choice options.
It therefore becomes an empirical question of whether item-level predictions from LLMs provide a good fit, if we aim at predicting the human data separately for each item, not as a condition-level average.

To address this question, we assessed item-level scores $\text{S}_{kl}$ from the text-davinci-003 instance of GPT-3.5 August 2023 \citep{BrownMann2020:Language-Models} for a text-based version of the reference game from Section~\ref{experiment-reference-games}.
The task description $x_{k}$ for item $I_{k}$ is a text supplied as prompt (see Appendix~\ref{sec:examples-items-llm} for an example).
The choice options are categorized as in the human experiment.\footnote{
  In the current set-up the response type ``distractor'' has two instantiations in the production condition. Since choice options are a single word in the production condition, for simplicity, we treat the union of the distractor words as a single option.}
Similar to the RSA model fit, we allow for random errors with probability $\epsilon_{C}$:
\begin{align*}
  P_{\text{item}}^{\text{LLM}}\left ( y_{kl} \mid C; \alpha_{c}, \epsilon_{c} \right )
  = (1- \epsilon_{c}) \  P_{\text{item}}^{\text{LLM}}\left ( y_{kl} \mid C; \alpha_{c}, \epsilon_{c} \right ) \ \ + \ \ \nicefrac{\epsilon_{c}}{3}   \,.
\end{align*}
With priors on parameters as for the RSA model described in Section~\ref{sec:model-pred-from}, the resulting likelihood function for item-level data with item-level predictions is:
\begin{align*}
 P^{\text{LLM}}_{\text{item}}(D_{c} \mid C; \alpha_{c}, \epsilon_{c}) = \prod_{k = 1}^{m} \text{Multinomial} \left ( D_{c}^{k}, \tuple{P^{\text{LLM}}_{\text{item}}(y_{kl} \mid C;  \alpha_{c}, \epsilon_{c})}_{1 \le l \le 3} \right )\,.
\end{align*}

\begin{figure}[t]
  \centering
  \includegraphics[width=\textwidth]{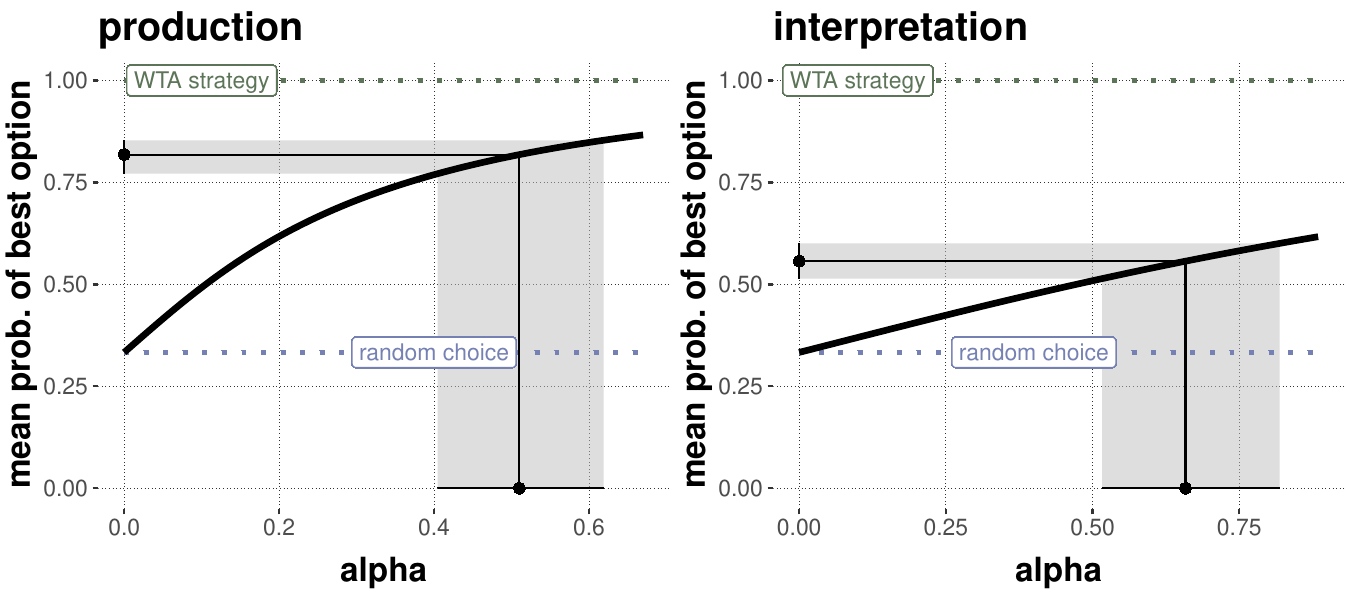}
  \caption{
    Predicted probability of highest-scoring answer category, averaged over items, for different values of softmax parameter $\alpha$ (black lines).
    The gray-shaded area indicates the posterior 95\% credible interval for $\alpha$, and the implied probabilistic prediction.
    For reference, the target probability under a random choice strategy and under the ``winner-takes-all'' (WTA) strategy are shown with dashed lines.
    For credible values of $\alpha$, the means of predicted probabilities for the target option are clearly distinct from the WTA strategy.
  }
  \label{fig:closeness-target-item-level}
\end{figure}

Summary statistics for samples from the posterior distribution over parameters are shown in Table~\ref{tab:sumStats} (rows 4 and 5).
Credible values of $\alpha$ parameters are comparatively low for the LLM-based item-level model, suggesting that the predicted scores have rather large differences, which have to be compensated for by the softmax link-function to adequately fit the data.
Figure~\ref{fig:closeness-target-item-level} shows the mean of the probabilities for the highest scoring answer category, averaged over all items, for different parameter values of $\alpha$, highlighting that for \textit{a posteriori} credible values of $\alpha$ (gray shaded areas), the predictions are clearly distinct from the predictions of a WTA strategy (which assigns probability 1 to the highest scoring option for each item).
This suggests that the WTA strategy, which is the special limiting case for $\alpha \rightarrow \infty$, provides a substantially worse fit for item-level choices than those provided by less extreme values of $\alpha$.

\begin{figure}[t]
  \centering

  \includegraphics[width = 0.9\linewidth]{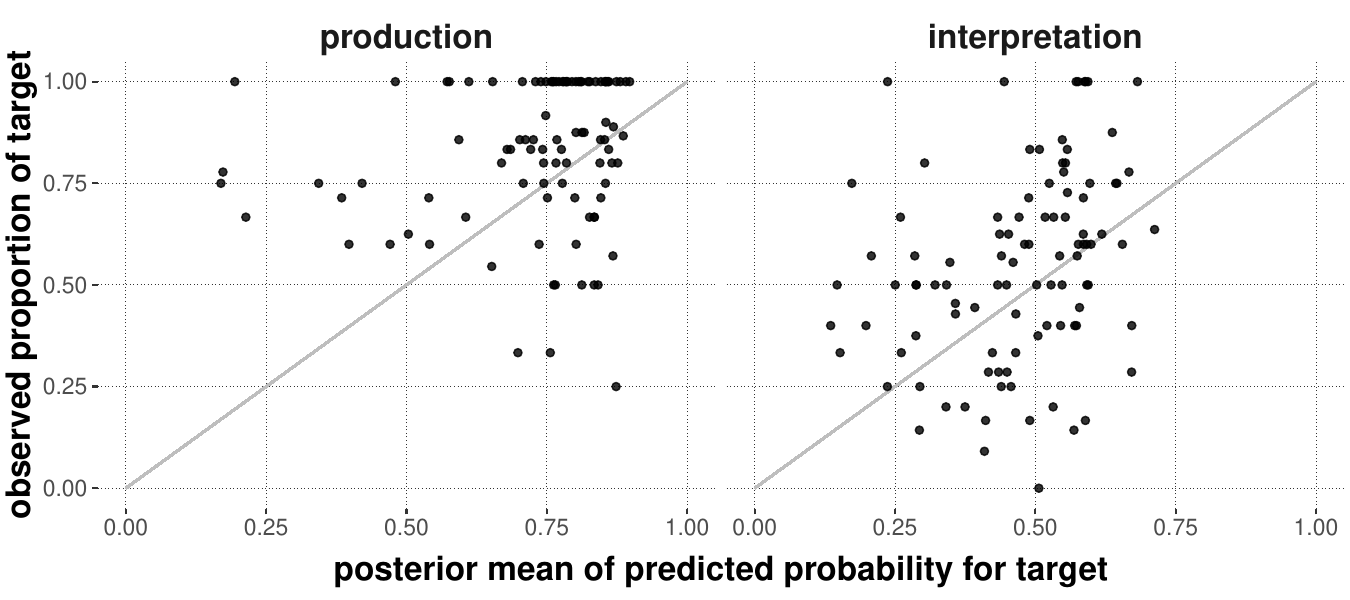}

  \caption{
    Item-level prediction-observation plot.
    Each dot represents an item.
    The $x$-coordinate represents the mean of the posterior prediction for the target choice probability for the given item.
    The $y$-coordinate represents the observed proportion of target choices for that item.
    The gray line (identity function) is the ideal prediction.
  }
  \label{fig:item-level-obs-pred}
\end{figure}

Sampling-based approximations of Bayesian posterior predictive $p$-values for the by-item analysis are very low (see Table~\ref{tab:sumStats}), suggesting that the unaggregated LLM scores are inadequate predictors of the human data.
To corroborate this conclusion, Figure~\ref{fig:item-level-obs-pred} shows that the item-level LLM-based model predicts variance which is not borne out by the human data.
Concretely, the plots show, for each condition and item, mean posterior estimates of the model's predicted probability of choosing the target option ($x$-axis), together with the observed proportion of target choices in the human data ($y$-axis).
There is ample variation in the model's predictions, especially visible in the production condition, owing to the fact that the item-level scores of the LLM sometimes clearly favor another option than the target choice.
So, the model itself predicts systematic variability at the item level.
The human data, too, show variability at the item-level, but there is no (visual) indication that the item-level variability predicted by the LLMs is borne out by the human data.
These results suggest that LLM-based probabilistic predictions may imply item-level variance that is not attested in the human data.
Put more strongly, a model that uses the most obvious item-level scores, derived from the predicted log-probabilities, to predict what human participants choose on a by-item level is ruled out by the current experimental data.

\section{Condition-level predictions}
\label{llm-predictions-for-reference-games}


\begin{figure}
  \centering
  \includegraphics[width=0.9\textwidth]{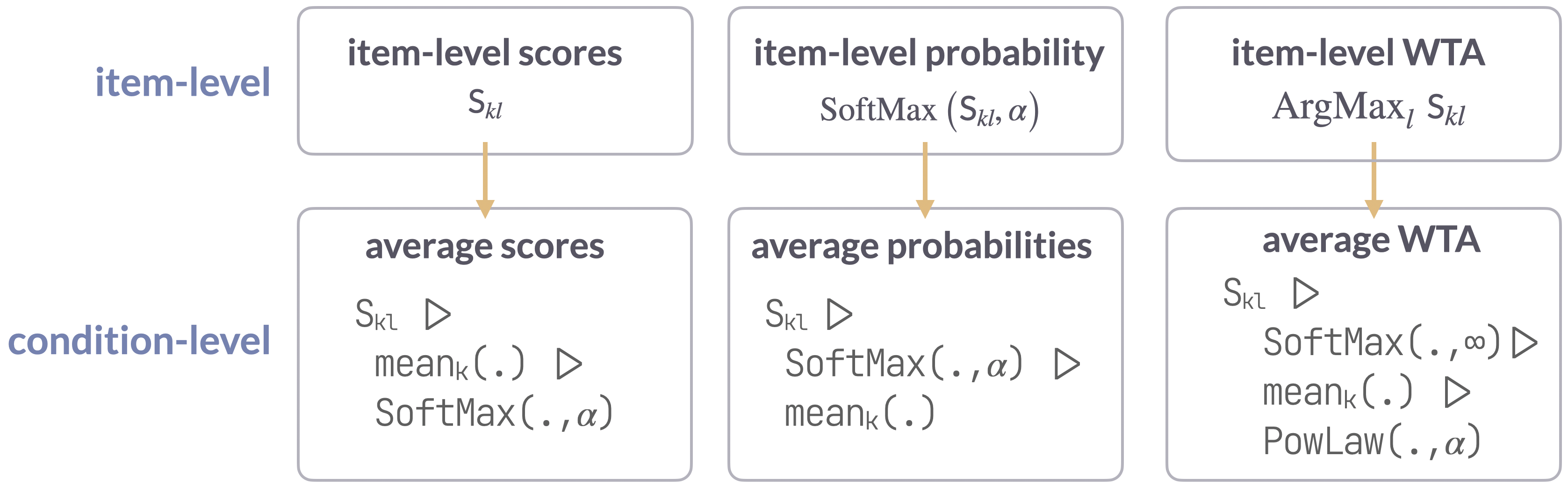}
  \caption{
    Schematic overview over three different methods of obtaining condition-level predictions by aggregating item-level information.
    Methods differ in the kind of item-level information they take into account, which entails differences in the order in which aggregation, transformation to probabilities and parameterized scaling occur.
  }
  \label{fig:measures-overview}
\end{figure}

While LLMs do not make conditional-level predictions as such, they can be derived from item-level scores $\text{S}_{kl}$ by averaging over all items belonging to the relevant condition.
There are many ways of averaging item-level information.
Figure~\ref{fig:measures-overview} shows three salient approaches, which differ in what the underlying item-level measure for aggregation is: the raw scores $\text{S}_{kl}$, the item-level probability derived from it (as used in Section~\ref{sec:item-level-pred}), or the predictions from the winner-takes-all (WTA) strategy commonly used in benchmark testing.


The \emph{average-scores predictor} first aggregates the item-level scores, and then transposes the averages into (scaled) probabilities using the usual parameterized softmax function:\footnote{
  The reported results average over the multi-set \(\set{I_1, \dots I_m}\) of items that occurred in the human experiment for condition $C$.
  By using a multi-set, which may contain a single item multiple times, we produce aggregate predictions for exactly the set of items that the participant group saw, which provides the most fitting counterpart to the human data.}
\begin{align*}
  & P_{\text{cond}}^{\text{SCR}}(R_{l} \mid C, \alpha_{c})
    \propto \expo \left [  \alpha_{c} \ \frac{1}{m} \ \sum_{k=1}^{m} \text{S}_{kl}  \right ] \,.
    \tag*{\textcolor{gray}{[average scores (narrow-scope aggregation)]}}
\end{align*}
The \emph{average-probabilities predictor} first transposes scores into probabilities with a parameterized softmax function, and only aggregates over items last:
\begin{align*}
  & P_{\text{cond}}^{\text{PRB}}(R_{l} \mid C, \alpha_{c})
    = \frac{1}{m} \ \sum_{k=1}^{m} P_{\text{item}}^{\text{LLM}}(y_{kl} \mid C, \alpha_{c}) \,.
    \tag*{\textcolor{gray}{[average probabilities (wide-scope aggregation)]}}
\end{align*}
Finally, the \emph{average-WTA predictor} considers the prediction of the WTA strategy (a softmax with $\alpha \rightarrow \infty$) as the basic item-level unit to aggregate over.
To add parameterized scaling to this method, a power-law transformation is a natural choice:
\begin{align*}
  & P_{\text{cond}}^{\text{WTA}}(R_{l} \mid C, \alpha_{c})
    \propto  \left [ \frac{1}{m} \ \sum_{k=1}^{m} P_{\text{item}}^{\text{WTA}}(y_{kl} \mid C) \right ]^{\alpha_{c}}\,,
    \tag*{\textcolor{gray}{[average WTA (intermediate-scope aggregation)]}}
\end{align*}
where $P_{\text{item}}^{\text{WTA}}(y_{kl} \mid C) = \lim_{\alpha \rightarrow \infty} P_{\text{item}}^{\text{LLM}}(y_{kl} \mid C, \alpha)$.
\bigskip

The \emph{average-scores} and \emph{average-probabilities} predictors are equivalent if there is only one item, in which case the prediction of the \emph{average-WTA} method is the special case of $\alpha \rightarrow \infty$.
For cases with more than one item, the predictions of the three predictors are not guaranteed to be the same.
Conceptually, the \emph{average-probabilities} and the \emph{average-WTA} predictors, but not the \emph{average-scores} predictor, are compatible with a picture in which condition-level predictions result from the actual predictions at the item level.
However, based on the results from the previous section, the item-level predictions for the WTA strategy are demonstrably incongruent with the human data.
In general, the average-probability and average-WTA predictors can lead to qualitatively different results for task-level accuracy (see Appendix~\ref{sec:aver-prob-vs}).

Using the same approach as for the RSA model in Section~\ref{sec:model-pred-from}, we can build likelihood functions for condition-level data around the three predictors introduced above.
With the same priors and methods used before, we obtain samples from the posterior over parameters and samples from the posterior predictive distributions.
Summary statistics for posteriors over model parameters are shown in Table~\ref{tab:sumStats}.
What is noteworthy is that for the \emph{average-WTA} model, the estimates of $\alpha_{c}$ in the production condition do not rule out, in fact lie close to, the special value $\alpha_{c}=1$, for which the power-law transformation is the identity function.
This means that just averaging WTA-responses at the item-level yields a reasonable predictor for the production data at the condition-level.
However, for the interpretation condition, the value $\alpha_{c}=1$ is clearly outside the range of credible parameter values, so that a simple recipe like ``always average WTA-responses without transformation'' is not a viable strategy for good condition-level predictions in general.
It is also worth noting that for the \emph{average-probability} model values of $\alpha_{c}$ of around five and above are virtually indistinguishable from the item-level predictions of the WTA strategy (see Figure~\ref{fig:closeness-target-item-level}).
This suggests that, for the production data, aggregation of item-level predictions from the WTA-strategy gives good condition-level predictions.

Figure~\ref{fig:refgame-counts} shows the summary statistics (means and 95\% credible intervals) for each model's posterior predictive distribution.
We find that only the theoretical model (RSA) and the \emph{average-WTA} model pass this ``visual posterior predictive check'' for both conditions; the other two models both overpredict the target choice rate and underpredict the competitor choice rate in the interpretation condition.
To corroborate the visual impression, Table~\ref{tab:sumStats} shows sample-based estimates of Bayesian posterior predictive $p$-values, using likelihood of the observed data as a test statistics.
Consequently, the results from Table~\ref{tab:sumStats} suggest that the \emph{average-scores} and the \emph{average-probabilities} models are able to reproduce the production data, but fail on the interpretation data; and that only the \emph{average-WTA} model does not fail to capture the data from both conditions.

These results tell us that not all ways of deriving condition-level predictions by averaging over item-level variation are equally good.
Some approaches clearly fail basic checks for statistical goodness-of-fit.
On the positive side, we also find that there is at least one model with predictors based on LLM-measures, namely the \emph{average-WTA} model, which is able to recover the patterns in the data.
In other words, there is a way of deriving predictor values for condition-level forced-choice probabilities from an LLM such that, when fed into a common linking function (here with parameters $\alpha$ for optimization and $\epsilon$ for random error), the human choice probability can be reconstructed faithfully in its entirety.
On the other hand, there is a stark contrast between the results of item-level and condition-level data.
The model that was able to properly fit the condition-level data builds on predictions for item-level choices that are clearly incompatible with the human item-level data.
The model seems to be ``right for the wrong reasons.''

\section{Generalizing to other LLM backends}
\label{sec:gener-other-llm}

\begin{figure}[t]
  \centering
  \includegraphics[width=0.9\textwidth]{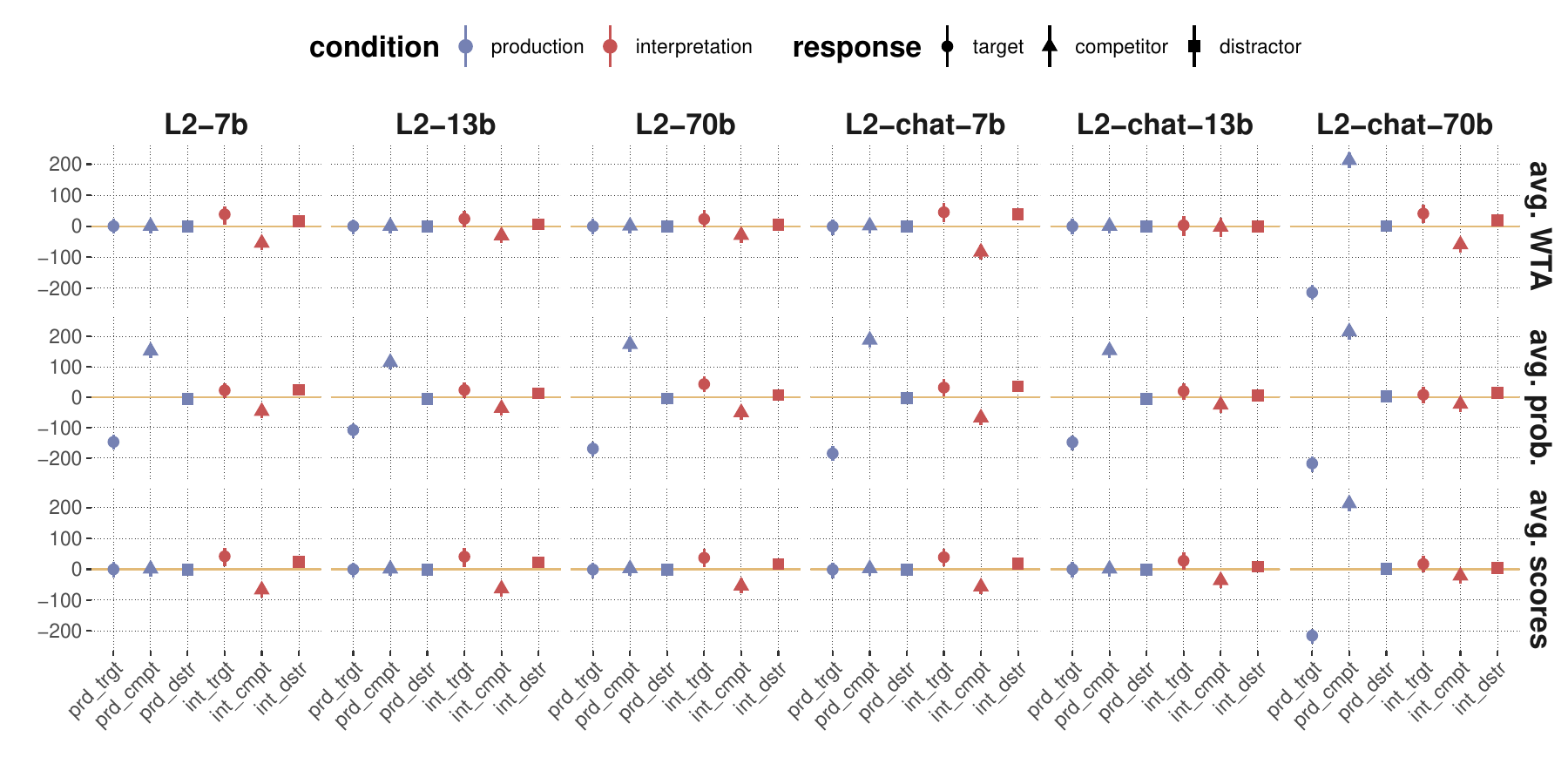}
  \caption{Visual posterior predictive checks on condition-level data for different models of the LLaMA family, paired with different aggregation strategies.}
  \label{fig:LLaMA-results-vPPC}
\end{figure}

All previous results where obtained for LMM predictions derived from GPT-3.5 davinci.
To investigate whether key results replicate for other LLMs and scales, we ran the same analyses also for scores derived from variants of LLaMA2 \citep{TouvronLavril2023:LLaMA:-Open-and}, in particular the 7, 13 and 70 billion parameter base models (marked as `base'), and the models of similar size fine-tuned on chat data (marked as `chat').
Table~\ref{tab:sumStats-LLaMA2} shows the relevant summary statistics and Bayesian posterior predictive $p$-values.
Figure~\ref{fig:LLaMA-results-vPPC} shows visual posterior predictive checks for the condition-level data.
The plots show the \emph{differences} between observed counts and the models' posterior predictions for expected counts for each condition and response option.
We find that the average-WTA predictor is consistently able to capture the data from the production condition, except for one of the six models (LLaMA2-chat 70B).
When the average-WTA predictor passes the model checking criterion for the production data, the posteriors for the power-law parameter $\alpha$ are credibly different from 1, thus suggesting that, contrary to the results for GPT-3.5 a mere ``WTA average'' is not a good prediction strategy for the production data for the LLaMA2-based models.
Looking at interpretation, the average-WTA predictor recovers the interpretation data for only one of the six models (LLaMA2-chat 13B, see also Table~\ref{tab:sumStats-LLaMA2}).
Interestingly, for LLaMA2-chat 13B also the average-probability predictor is able to recover the interpretation data, unlike for the GPT-3.5 model.
Finally, for item-level data, as for GPT-3.5, no LLaMA2-based model captured the item-level variance observed in the human data.
In sum, we find general support for the previous conclusions that the average-WTA predictor is most successful in capturing the condition-level data, and therefore that, where models manage to recover the condition-level data, they seem to be ``right for the wrong reasons.''
Additionally, we also find variability in which method of aggregation works well with which LLM as scoring model.

\begin{table}[]
\centering
\begin{tabular}{cllllcrlcrlc}
  \toprule
  &&&&&& $\alpha$ &&& $\epsilon$ & \\ \cmidrule(r){6-8} \cmidrule(l){9-11}
  & model & data & method & condition & |95\% & mean\ & 95\%| & |95\% & mean\ & 95\%| & Bpppv \\
  \midrule

  \textcolor{shimmer-main}{$\times$} & L2-base-7b    & item  & ---         & prd. & 1.09  & 1.30   & 1.51   & 0.00 & 0.02 & 0.06 & 0.00 \\
  \textcolor{shimmer-main}{$\times$} & L2-base-7b    & item  & ---         & int. & 1.90  & 2.52   & 3.13   & 0.00 & 0.04 & 0.12 & 0.00 \\
  \textcolor{fern-main}{$\checkmark$} & L2-base-7b    & cond. & avg. scores & prd. & 9.65  & 11.50  & 13.62  & 0.08 & 0.12 & 0.16 & 0.48 \\
  \textcolor{shimmer-main}{$\times$} & L2-base-7b    & cond. & avg. scores & int. & 6.80  & 7.96   & 9.14   & 0.00 & 0.02 & 0.05 & 0.00 \\
  \textcolor{shimmer-main}{$\times$} & L2-base-7b    & cond. & avg. prob.  & prd. & 4.98  & 16.44  & 33.06  & 0.06 & 0.10 & 0.13 & 0.00 \\
  \textcolor{shimmer-main}{$\times$} & L2-base-7b    & cond. & avg. prob.  & int. & 20.06 & 79.30  & 163.94 & 0.00 & 0.01 & 0.04 & 0.00 \\
  \textcolor{fern-main}{$\checkmark$} & L2-base-7b    & cond. & avg. WTA    & prd. & 4.17  & 4.96   & 5.85   & 0.07 & 0.12 & 0.16 & 0.48 \\
  \textcolor{shimmer-main}{$\times$} & L2-base-7b    & cond. & avg. WTA    & int. & 0.87  & 1.02   & 1.18   & 0.00 & 0.02 & 0.05 & 0.00 \\ \addlinespace[0.5em]
  \textcolor{shimmer-main}{$\times$} & L2-base-13b   & item  & ---         & prd. & 1.37  & 1.91   & 2.47   & 0.00 & 0.07 & 0.12 & 0.00 \\
  \textcolor{shimmer-main}{$\times$} & L2-base-13b   & item  & ---         & int. & 2.13  & 2.90   & 3.69   & 0.00 & 0.05 & 0.14 & 0.00 \\
  \textcolor{fern-main}{$\checkmark$} & L2-base-13b   & cond. & avg. scores & prd. & 13.73 & 16.32  & 19.10  & 0.08 & 0.12 & 0.16 & 0.50 \\
  \textcolor{shimmer-main}{$\times$} & L2-base-13b   & cond. & avg. scores & int. & 7.73  & 9.11   & 10.42  & 0.00 & 0.02 & 0.05 & 0.00 \\
  \textcolor{shimmer-main}{$\times$} & L2-base-13b   & cond. & avg. prob.  & prd. & 69.84 & 185.80 & 360.44 & 0.06 & 0.09 & 0.13 & 0.00 \\
  \textcolor{shimmer-main}{$\times$} & L2-base-13b   & cond. & avg. prob.  & int. & 13.63 & 39.72  & 88.28  & 0.00 & 0.02 & 0.05 & 0.00 \\
  \textcolor{fern-main}{$\checkmark$} & L2-base-13b   & cond. & avg. WTA    & prd. & 2.28  & 2.70   & 3.18   & 0.08 & 0.12 & 0.16 & 0.51 \\
  \textcolor{shimmer-main}{$\times$} & L2-base-13b   & cond. & avg. WTA    & int. & 0.78  & 0.94   & 1.12   & 0.00 & 0.03 & 0.07 & 0.03 \\ \addlinespace[0.5em]
  \textcolor{shimmer-main}{$\times$} & L2-base-70b   & item  & ---         & prd. & 1.53  & 1.86   & 2.22   & 0.00 & 0.03 & 0.07 & 0.00 \\
  \textcolor{shimmer-main}{$\times$} & L2-base-70b   & item  & ---         & int. & 3.20  & 4.01   & 4.74   & 0.00 & 0.03 & 0.08 & 0.00 \\
  \textcolor{fern-main}{$\checkmark$} & L2-base-70b   & cond. & avg. scores & prd. & 44.57 & 53.17  & 62.44  & 0.08 & 0.12 & 0.17 & 0.48 \\
  \textcolor{shimmer-main}{$\times$} & L2-base-70b   & cond. & avg. scores & int. & 6.86  & 8.14   & 9.32   & 0.00 & 0.02 & 0.05 & 0.00 \\
  \textcolor{shimmer-main}{$\times$} & L2-base-70b   & cond. & avg. prob.  & prd. & 10.51 & 26.59  & 45.73  & 0.07 & 0.10 & 0.14 & 0.00 \\
  \textcolor{shimmer-main}{$\times$} & L2-base-70b   & cond. & avg. prob.  & int. & 10.86 & 23.63  & 42.16  & 0.00 & 0.03 & 0.09 & 0.00 \\
  \textcolor{fern-main}{$\checkmark$} & L2-base-70b   & cond. & avg. WTA    & prd. & 5.88  & 7.03   & 8.19   & 0.08 & 0.12 & 0.17 & 0.50 \\
  \textcolor{shimmer-main}{$\times$} & L2-base-70b   & cond. & avg. WTA    & int. & 0.55  & 0.68   & 0.81   & 0.00 & 0.03 & 0.07 & 0.04 \\ \addlinespace[0.5em]
  \textcolor{shimmer-main}{$\times$} & L2-chat-7b  & item  & ---         & prd. & 0.34  & 0.41   & 0.49   & 0.00 & 0.03 & 0.07 & 0.00 \\
  \textcolor{shimmer-main}{$\times$} & L2-chat-7b  & item  & ---         & int. & 0.79  & 1.05   & 1.34   & 0.00 & 0.05 & 0.15 & 0.00 \\
  \textcolor{fern-main}{$\checkmark$} & L2-chat-7b  & cond. & avg. scores & prd. & 58.47 & 69.64  & 81.79  & 0.08 & 0.12 & 0.16 & 0.48 \\
  \textcolor{shimmer-main}{$\times$} & L2-chat-7b  & cond. & avg. scores & int. & 2.03  & 2.37   & 2.73   & 0.00 & 0.02 & 0.05 & 0.00 \\
  \textcolor{shimmer-main}{$\times$} & L2-chat-7b  & cond. & avg. prob.  & prd. & 8.05  & 27.26  & 55.30  & 0.07 & 0.11 & 0.15 & 0.00 \\
  \textcolor{shimmer-main}{$\times$} & L2-chat-7b  & cond. & avg. prob.  & int. & 2.86  & 19.24  & 51.41  & 0.00 & 0.01 & 0.04 & 0.00 \\
  \textcolor{fern-main}{$\checkmark$} & L2-chat-7b  & cond. & avg. WTA    & prd. & 12.18 & 14.70  & 17.21  & 0.08 & 0.12 & 0.16 & 0.50 \\
  \textcolor{shimmer-main}{$\times$} & L2-chat-7b  & cond. & avg. WTA    & int. & 0.86  & 1.01   & 1.14   & 0.00 & 0.02 & 0.05 & 0.00 \\ \addlinespace[0.5em]
  \textcolor{shimmer-main}{$\times$} & L2-chat-13b & item  & ---         & prd. & 0.36  & 0.44   & 0.53   & 0.00 & 0.03 & 0.08 & 0.00 \\
  \textcolor{shimmer-main}{$\times$} & L2-chat-13b & item  & ---         & int. & 0.68  & 0.90   & 1.11   & 0.00 & 0.05 & 0.13 & 0.00 \\
  \textcolor{fern-main}{$\checkmark$} & L2-chat-13b & cond. & avg. scores & prd. & 5.56  & 6.63   & 7.76   & 0.08 & 0.12 & 0.16 & 0.49 \\
  \textcolor{shimmer-main}{$\times$} & L2-chat-13b & cond. & avg. scores & int. & 1.70  & 2.05   & 2.40   & 0.00 & 0.02 & 0.06 & 0.01 \\
  \textcolor{shimmer-main}{$\times$} & L2-chat-13b & cond. & avg. prob.  & prd. & 3.90  & 22.25  & 74.12  & 0.06 & 0.10 & 0.14 & 0.00 \\
  \textcolor{fern-main}{$\checkmark$} & L2-chat-13b & cond. & avg. prob.  & int. & 2.85  & 4.70   & 7.01   & 0.00 & 0.03 & 0.07 & 0.08 \\
  \textcolor{fern-main}{$\checkmark$} & L2-chat-13b & cond. & avg. WTA    & prd. & 4.28  & 5.11   & 5.99   & 0.07 & 0.12 & 0.16 & 0.49 \\
  \textcolor{fern-main}{$\checkmark$} & L2-chat-13b & cond. & avg. WTA    & int. & 0.15  & 0.32   & 0.51   & 0.02 & 0.11 & 0.18 & 0.50 \\ \addlinespace[0.5em]
  \textcolor{shimmer-main}{$\times$} & L2-chat-70b & item  & ---         & prd. & 0.28  & 0.35   & 0.42   & 0.00 & 0.03 & 0.08 & 0.00 \\
  \textcolor{shimmer-main}{$\times$} & L2-chat-70b & item  & ---         & int. & 1.03  & 1.24   & 1.47   & 0.00 & 0.03 & 0.08 & 0.00 \\
  \textcolor{shimmer-main}{$\times$} & L2-chat-70b & cond. & avg. scores & prd. & 0.40  & 0.55   & 0.74   & 0.00 & 0.04 & 0.10 & 0.00 \\
  \textcolor{fern-main}{$\checkmark$} & L2-chat-70b & cond. & avg. scores & int. & 1.91  & 2.37   & 2.89   & 0.00 & 0.03 & 0.08 & 0.16 \\
  \textcolor{shimmer-main}{$\times$} & L2-chat-70b & cond. & avg. prob.  & prd. & 0.43  & 0.66   & 0.96   & 0.00 & 0.04 & 0.10 & 0.00 \\
  \textcolor{shimmer-main}{$\times$} & L2-chat-70b & cond. & avg. prob.  & int. & 2.82  & 5.21   & 8.43   & 0.00 & 0.01 & 0.04 & 0.02 \\
  \textcolor{shimmer-main}{$\times$} & L2-chat-70b & cond. & avg. WTA    & prd. & 0.14  & 0.21   & 0.35   & 0.00 & 0.06 & 0.13 & 0.00 \\
  \textcolor{shimmer-main}{$\times$} & L2-chat-70b & cond. & avg. WTA    & int. & 0.85  & 1.00   & 1.14   & 0.00 & 0.02 & 0.05 & 0.00 \\

   \bottomrule \\
\end{tabular}
\caption{
  Summary statistics for models based on LLaMA2 variants.
  Information shown is the same as in Table~\ref{tab:sumStats}.
}
\label{tab:sumStats-LLaMA2}
\end{table}

\section{Conclusion}
\label{sec:conclusion}

While the common practice in evaluating the capabilities of LLMs is based on accuracy averaged over large collections of data, this work took the alternative route to explore what we learn if we subjected LLMs to the same routines and strong demands on distributional quality of fit to human data as we normally do for statistical or probabilistic cognitive models.
Knowledge of the adequacy of LLMs' full distributional predictions for simple tasks that require human-like judgement or decisions is important to gauge in how far LLMs can be trusted to provide such information in applications such as hybrid, neuro-symbolic models \citep[e.g.,][]{GarneloShanahan2019:Reconciling-dee,LewTessler2020:Leveraging-Unst,CreswellShanahan2022:Selection-Infer,Frank2023:Large-language-}.
A main contribution of this paper is methodological, showing how statistical model criticism can be used for LLMs in the first place, and, more specifically, how it can be insightful in the detailed assessment of how LLMs might or might not be able to predict human data at the degree of accuracy that we would normally aspire for when dealing with explanatory statistical models in cognitive science.
Applying this comparative approach to a minimal, but non-trivial data set, we find that the LLM predictions on a per-item level predict variance that is not attested in the human data.
From several candidate predictor measures for aggregate condition-level data, only one was not refuted by the human data (at least for the GPT-3.5 model), but this was one that relied on the empirically implausible WTA-strategy at the item-level, incidentally the same strategy that is commonly used in accuracy-based benchmark testing \citep{srivastava2023-BIGbench}.

\paragraph{Explanatory power.}
A basic observation brought to the foreground by our approach is that LLMs' atomic predictions are for individual items and that some aggregation method is needed to derive more abstract, condition- or task-level predictions.
This is one sense in which LLMs may be felt to be less, or not at all, explanatory.
They do not offer, at least not directly, a human-comprehensible compression of reality into a \emph{kind} of response pattern, over and beyond making a prediction for each particular situation.
As this kind of compression is arguably important for a sense of understanding \citep{Dellsen2020:Beyond-Explanat,Grimm2021:Understanding}, the direct comparison of LLMs with common practices in experimental psychology and with probabilistic cognitive models, provides an interesting perspective on why LLMs are often felt to be lacking in explanatory power.

This perspective on the explanatory role of LLMs goes beyond the factors of performance, indirect support and parsimony identified by \citet{SortNoopDeemtervan-Deemter2023:Dimensions-of-E}.
It is also subtly different from considerations of an LLM's ability to generalize \citep{HupkesDankers2020:Compositionalit}.
It rather suggests that \emph{transferability} is a dimension to ``explanatory power'' that is important as well.
Imagine that models $M$ and $M'$ have been designed for and trained on data from a situation $S$, but need to be applied to a different situation $T$.
Assume that for model $M$ the only way to make predictions for $T$ is to collect data pertaining to $T$, and either retrain or fine-tune the model.
In contrast, model $M'$ can make predictions for $T$ without novel data collection by recognizing a meaningful difference between $S$ and $T$ and consequently manually changing parameter values or model-internal mechanics to accommodate for this change.
In that case, model $M'$ would be more transferable than model $M$.
For example, if we change the experimental setting for a reference game to consist of data from a special population, such as very young children or language-impaired adults, a potentially reasonable architectural change to a model like the RSA model is to consider differences in the sets of alternative utterances for the speaker \citep[e.g.][]{Noveck2001:When-Children-a}.
Even though this is only a vague explication of a notion of transferability, it suffices to corroborate the intuition that probabilistic cognitive models like the Rational Speech Act model, which are designed to operate at a higher level of conceptual abstraction, will often appear more transferable than models like LLMs, which make predictions not for \emph{kinds} of situations but for \emph{particular} situations.
Whether any given model's transfer-ability is correct, is an orthogonal empirical question.
It is also an empirical question, exactly to which extent and in which areas LLMs are (not) transferable, especially if we consider prompting a transfer strategy \citep{LiuLiu2022:Generated-Knowl,XieRaghunathan2022:An-Explanation-}.
In any case, the comparison between LLMs and other statistical or probabilistic cognitive models started here suggests systematic research into the transfer-ability and explanatory value of LLMs, e.g., by prompting strategies that are empirically insightful or by their use in composite neuro-symbolic models that implement theoretically-meaningful conceptual differences between types of situations.


\paragraph{Variability in predictions and performance measures.}
Despite being anchored in a small, but detailed case-study, the fact that different plausible methods of aggregating item-level information led to condition-level predictions of variable quality is worrisome.
The wide-spread reliance on a winner-takes-all strategy might be inconsistent with the actual use practices of LLMs, which may not always rely on a temperature-zero sampling strategy (see Appendix~\ref{sec:aver-prob-vs}).
In conclusion, research on LLMs should systematically study the conceptual and empirical consequences of seemingly minor decisions in evaluation or application settings.
Variability of performance measures also comes with a well-known risk for robust and reproducible science \citep[c.f.][]{HuLevy2023:Prompt-based-me,TsvilodubWang2024:Predictions-fro}.
The more researcher degrees of freedom there are, the higher the risk of false results, even in the absence of intentions to mislead \citep[e.g.][]{Ioannidis2005:Why-Most-Publis,Chambers2017:The-Seven-Deadl}.
Testing LLMs as predictors in statistical models should raise awareness for the issue of robust research methods also in NLP \citep{WielingRawee2018:Reproducibility}.

The work presented here also contains considerable researcher degrees of freedom.
We only considered one out of several conceivable ways of carving a parameterized likelihood function from LLM-based predictors.
If future work would contribute to systematically exploring these, the main goal of this paper would have been met, which is to raise awareness for the possibility, perhaps even necessity, to scrutinize LLMs at the same level of rigorous detail as other models in cognitive science.

\paragraph{Human-like predictions from LLM.}
The question after the human-likeness of quantitative LLM-derived information matters for applications which use numerical scores to rank or weigh options \citep[e.g.,][]{ParkOBrien2023:Generative-Agen,ZhangLehman2023:OMNI:-Open-ende}.
Moreover, to the extent that LLMs are used as parts of explanatory ``neuro-symbolic models'' of information processing \citep{GarcezLamb2020:Neurosymbolic-A}, understanding whether and how LLMs might yield full-fledged distributional predictions is important, e.g., to explore their integration into probabilistic (cognitive) models \citep[c.f.,][]{LewTessler2020:Leveraging-Unst,WongGrand2023:From-Word-Model,Frank2023:Large-language-}.
Based on the data set and the detailed analyses conducted here, it seems not infeasible to use numerical predictions from LLMs as part of predictive probabilistic models.
But, ideally, low-level prompt variation, such as from order of presentation or similar ``nuisance variables,'' should be averaged out or taken into account in some way or other, as we do not understand what causes this variation in the predictions of models.
Further research is necessary that investigates when exactly this variation accords with empirically observed patterns.
In sum, we conjecture that using LLM predictors for probabilistic predictions, such as in a neuro-symbolic model, might be possible if embedded in the proper link functions and if item-level variation is taken into account.
This, however, entails that each LLM component in a hybrid model should be independently tested against at least a modestly sized empirical data set.

\paragraph{No substitute for human subjects.}
Looking at LLM-derived predictions for human data from the perspective of cognitive modeling highlights the fact that LLMs predict item-level variation, but not subject-level variation, which is common in human data.
We may consider variation introduced via softmax / temperature as analogous to between-human variation, but this likely falls short of reality, where pronounced differences in answer behavior may surface.
For the particular case of pragmatic language use, prior research has shown that individual participants have markedly different behavioral profiles, often consistently behaving like literal language users or more sophisticated language users \citep[e.g.,][]{NieuwlandDitman2010:On-the-incremen,FrankeDegen2015:Reasoning-in-Re,SpychalskaKontinen2016:Investigating-s}.
It is an open question whether predictions from LLMs reflect the same kind of variation \citep[][]{AherArriaga2023:Using-large-lan,SalewskiAlaniz2023:In-Context-Impe}.
The results presented here recommend skepticism.
We therefore side with the cautious voices that do not recommend replacing human participants with LLMs in psychological research \citep{DillionTandon2023:Can-AI-language,HardingDAlessandro2023:AI-language-mod}.
In contrast, this points to an important challenge for future LLM research.
The fact that aggregate predictions can track aggregate human behaviour means that the variance on both sides is washed out to achieve a similar result.
This raises the issue of finding the systematic differences between LLMs and human answerers.
The task then would be to find cases where the differences do \emph{not} wash out and ask: What if anything do these cases share?

\paragraph{Limitations and follow-up work.}
The scope of our experimental investigation was deliberately small but perspicuous.
Focusing on the methodological contributions and details in performance assessments, we investigated a minimal non-trivial case study where we could triangulate LLMs, probabilistic cognitive models and human data.
This work may therefore serve as starting point for a wider investigation of more complex data sets and case studies in which LLMs are analyzed as and directly compared to explanatory PCMs.

\section*{Acknowledgements}
Thanks to Robert D. Hawkins and Thomas Brochhagen for insightful comments.
We gratefully acknowledge support by the state of Baden-Württemberg, Germany, through the computing resources provided by bwHPC and the German Research Foundation (DFG) through grant INST 35/1597-1 FUGG.
MF is a member of the Machine Learning Cluster of Excellence, EXC number 2064/1 – Project number 39072764.

\printbibliography[heading=bibintoc]

\newpage
\appendix

\section{Screenshots from the online experiment with human participants}
\label{sec:scre-from-online}

\begin{figure}[H]
  \centering
  \includegraphics[width = 0.8\textwidth]{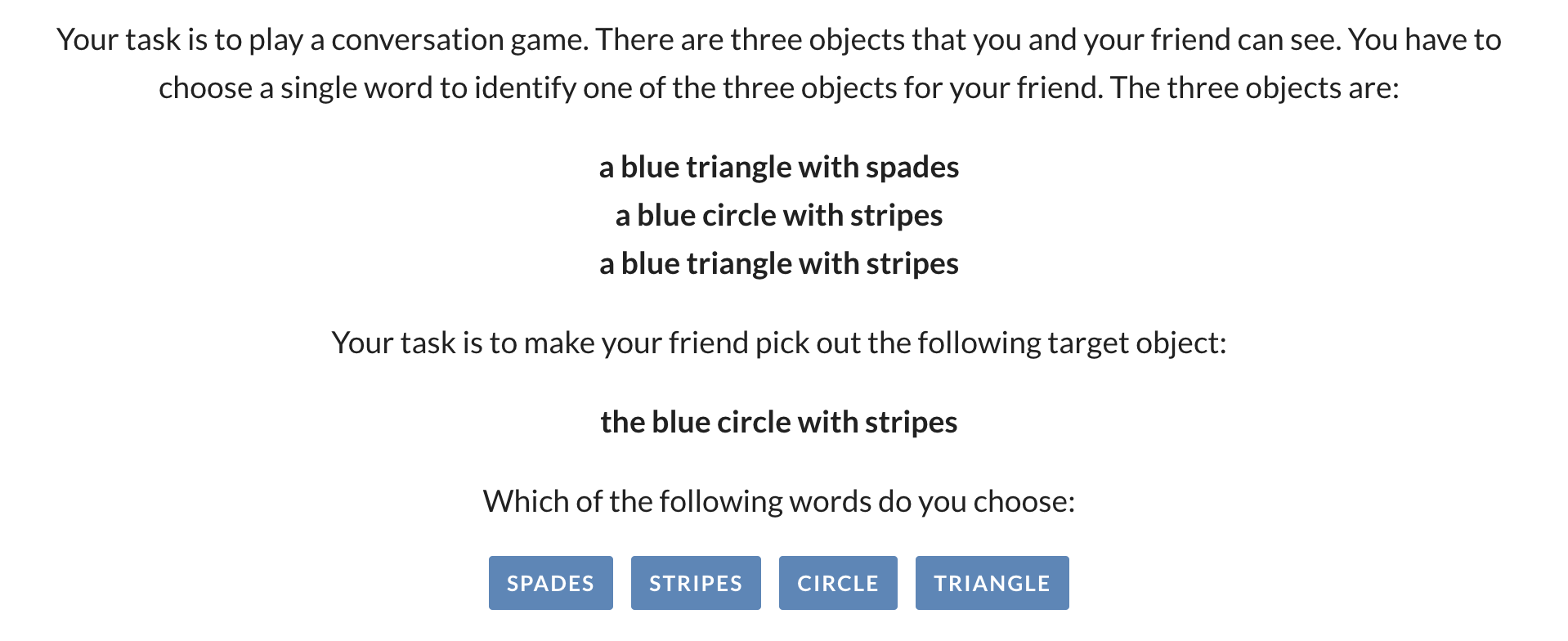}

  \caption{Screen shot from a production trial of the online experiment.}
  \label{fig:refgame-screenshot-production}
\end{figure}

\begin{figure}[H]
  \centering
  \includegraphics[width = 0.8\textwidth]{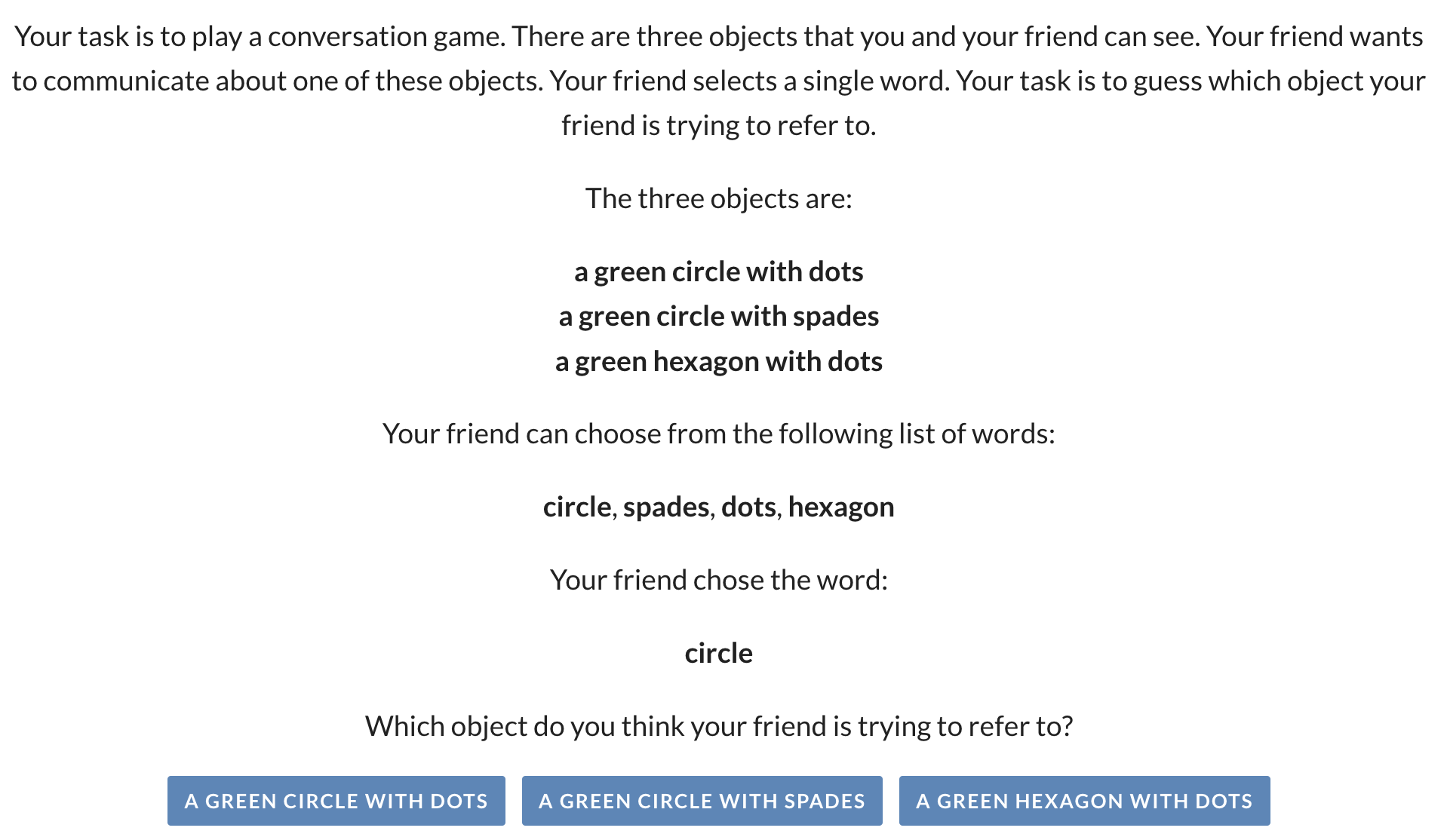}

  \caption{Screen shot from an interpretation trial of the online experiment.}
  \label{fig:refgame-screenshot-interpretation}
\end{figure}

Figure~\ref{fig:refgame-screenshot-production} shows a trial from the production condition, Figure~\ref{fig:refgame-screenshot-interpretation} one for the interpretation condition of the online experiment.

\section{Example item for the LLM experiment}
\label{sec:examples-items-llm}

The text-based input for the LLM predictions mirrors the text in the human experiment, except that the LLM input also lists the set of all available choice options (which for the human experiment is unnecessary since this information is given by the buttons for the forced-choice selection).
For example, the task description $T_{k}$ for the item that corresponds to the production trial shown in Figure~\ref{fig:refgame-screenshot-production} is shown below (the actual input has no line breaks in the first paragraph):

\begin{verbatim}
Your task is to play a conversation game. There are three objects that
you and your friend can see. You have to choose a single word to identify
one of the three objects for your friend.

The three objects are:

a blue triangle with spades
a blue circle with stripes
a blue triangle with stripes

Your task is to make your friend pick out the following target object:

the blue circle with stripes

Which of the following words would you choose:

spades
stripes
circle
triangle

Your answer:

I would choose the word
\end{verbatim}

\section{Excursion: Accuracy scores from average-probability vs.~average-WTA predictors}
\label{sec:aver-prob-vs}

Standard benchmark testing looks at \emph{task accuracy}, defined as the probability of selecting the ``gold standard'' response, usually based on ``winner-takes-all'' (WTA) selection of the highest scoring option.
We can generalize this and define the \emph{softmax-based accuracy} as the average probability of choosing the designated target response option $R_{1}$ for the \emph{condition-level soft-max prediction} (notation as defined in main text):
\begin{align*}
  P_{\text{cond}}^{\text{SM}}\left(R_{1} \right) = \frac{1}{m} \sum_{k = 1}^{m} P_{\text{item}}^{\text{SM}} \left(y_{k1} \right)\,.
\end{align*}
The WTA-based accuracy (the standard measure in most benchmark testing) is the special case of $\alpha \rightarrow\infty$.

The WTA-based accuracy can differ qualitatively from the more general softmax-based accuracy, as shown by the following example.
\begin{quote}
  \textbf{Example:}
  Imagine that there are two options, and that the target option's score is a small $\epsilon$ higher in 80\% of the task's items, and otherwise lower.
  The WTA-based accuracy is 0.8.
  This number is useful as a performance measure for applications in which the LLM is used in exactly the way the WTA strategy describes, e.g., any implementation which is outcome equivalent to greedy decoding with rejection sampling on a domain that contains only the available options.
  For such a case, it never matters how much worse the goal answer is scored in the 20\% of the cases where it is not the maximum.
  As only the best option will be chosen, that information is irrelevant.
  But if an application uses anything other than greedy-like responses, the accuracy score of 0.8 may be misleading.
  If the remaining 20\% of the items are such that the non-goal option is almost infinitely better, it would be chosen under a pure sampling strategy, where $\alpha = 1$, with virtual certainty, so the softmax-based accuracy would be around 0.4.\footnote{The probability of the target option in the 80\% of items where the goal answer is slightly better is 0.5 in the limit of $\epsilon \rightarrow 0$, and it is virtually 0 in the remaining 20\% of the cases. This gives an expected rate of: $\nicefrac{4}{5} \ \nicefrac{1}{2} + \nicefrac{1}{5} \ 0 = \nicefrac{2}{5}$.}
\end{quote}
The example shows that differences between accuracy measures depend on the variation in item-level scores, in particular on the relation between score-ordering and score-differences.
Note that the example holds equally if numbers for the two options are reversed, so that there is no way of saying which of the two measures of accuracy would generally be more favorable for selecting the target option.

The upshot of these considerations is that the standard practice of WTA-based performance assessment for LLMs gives false, or at least misleading or inaccurate results, whenever not all downstream applications use a greedy-like sampling strategy (which is almost certainly the case), and there is variability in item-level predictions (which may or may not be the case, depending on the domain of application).

\end{document}